\pdfoutput=1
\documentclass[twoside,american,english]{IEEEtran}
\usepackage[T1]{fontenc}
\usepackage[latin9]{inputenc}
\usepackage[letterpaper]{geometry}
\usepackage{xcolor}
\usepackage{array}
\usepackage{float}
\usepackage{units}
\usepackage{mathtools}
\usepackage{multirow}
\usepackage{amsmath}
\usepackage{amsthm}
\usepackage{amssymb}
\usepackage{graphicx}
\usepackage{wasysym}

\makeatletter

\newcommand{\noun}[1]{\textsc{#1}}
\providecommand{\tabularnewline}{\\}

\floatstyle{ruled}
\newfloat{algorithm}{tbp}{loa}
\providecommand{\algorithmname}{Algorithm}
\floatname{algorithm}{\protect\algorithmname}

\theoremstyle{plain}
\ifx\thechapter\undefined
	\newtheorem{thm}{\protect\theoremname}
\else
	\newtheorem{thm}{\protect\theoremname}[chapter]
\fi
\theoremstyle{definition}
\newtheorem{example}[thm]{\protect\examplename}
\theoremstyle{plain}
\newtheorem{prop}[thm]{\protect\propositionname}
\theoremstyle{definition}
\newtheorem{defn}[thm]{\protect\definitionname}

\usepackage{graphicx}
\usepackage{import}
\usepackage{makecell}
\graphicspath{{images/}}

\usepackage{anyfontsize} 
\usepackage[font=footnotesize]{caption} 
\usepackage{url}

\usepackage{makecell} 

\usepackage{algorithm,algpseudocode}
\algnewcommand{\LineComment}[1]{\State \(\triangleright\) #1}
\algnewcommand\algorithmicnot{\textbf{not}}
\algdef{SE}[IF]{IfNot}{EndIf}[1]{\algorithmicif\ \algorithmicnot\ #1\ \algorithmicthen}{\algorithmicend\ \algorithmicif}%

\usepackage{tikz}
\newcommand*\circled[1]{\tikz[baseline=(char.base)]{ \node[shape=circle,draw,inner sep=1pt, fill=black, text=white] (char) {#1};}}

\ifdefined\showcaptionsetup
 \PassOptionsToPackage{caption=false}{subfig}
\fi
\usepackage{subfig}
\makeatother

\usepackage{babel}
\addto\captionsamerican{\renewcommand{\algorithmname}{Algorithm}}
\addto\captionsamerican{\renewcommand{\definitionname}{Definition}}
\addto\captionsamerican{\renewcommand{\examplename}{Example}}
\addto\captionsamerican{\renewcommand{\propositionname}{Proposition}}
\addto\captionsamerican{\renewcommand{\theoremname}{Theorem}}
\addto\captionsenglish{\renewcommand{\definitionname}{Definition}}
\addto\captionsenglish{\renewcommand{\examplename}{Example}}
\addto\captionsenglish{\renewcommand{\propositionname}{Proposition}}
\addto\captionsenglish{\renewcommand{\theoremname}{Theorem}}
\providecommand{\definitionname}{Definition}
\providecommand{\examplename}{Example}
\providecommand{\propositionname}{Proposition}
\providecommand{\theoremname}{Theorem}

\begin{document}
\global\long\def\dq#1{\underline{\boldsymbol{#1}}}%

\global\long\def\quat#1{\boldsymbol{#1}}%

\global\long\def\mymatrix#1{\boldsymbol{#1}}%

\global\long\def\myvec#1{\boldsymbol{#1}}%

\global\long\def\mapvec#1{\boldsymbol{#1}}%

\global\long\def\dualvector#1{\underline{\boldsymbol{#1}}}%

\global\long\def\dual{\varepsilon}%

\global\long\def\dotproduct#1{\langle#1\rangle}%

\global\long\def\norm#1{\left\Vert #1\right\Vert }%

\global\long\def\mydual#1{\underline{#1}}%

\global\long\def\hamilton#1#2{\overset{#1}{\operatorname{\mymatrix H}}\left(#2\right)}%

\global\long\def\hamiquat#1#2{\overset{#1}{\operatorname{\mymatrix H}}_{4}\left(#2\right)}%

\global\long\def\hami#1{\overset{#1}{\operatorname{\mymatrix H}}}%

\global\long\def\tplus{\dq{{\cal T}}}%

\global\long\def\getp#1{\operatorname{\mathcal{P}}\left(#1\right)}%

\global\long\def\getd#1{\operatorname{\mathcal{D}}\left(#1\right)}%

\global\long\def\swap#1{\text{swap}\left(#1\right)}%

\global\long\def\imi{\hat{\imath}}%

\global\long\def\imj{\hat{\jmath}}%

\global\long\def\imk{\hat{k}}%

\global\long\def\real#1{\operatorname{\mathrm{Re}}\left(#1\right)}%

\global\long\def\imag#1{\operatorname{\mathrm{Im}}\left(#1\right)}%

\global\long\def\imvec{\boldsymbol{\imath}}%

\global\long\def\vector{\operatorname{vec}}%

\global\long\def\mathpzc#1{\fontmathpzc{#1}}%

\global\long\def\cost#1#2{\underset{\text{#2}}{\operatorname{\text{C}}}\left(\ensuremath{#1}\right)}%

\global\long\def\diag#1{\operatorname{diag}\left(#1\right)}%

\global\long\def\frame#1{\mathcal{F}_{#1}}%

\global\long\def\ad#1#2{\text{Ad}\left(#1\right)#2}%

\global\long\def\spin{\text{Spin}(3)}%

\global\long\def\spinr{\text{Spin}(3){\ltimes}\mathbb{R}^{3}}%

\global\long\def\f#1{\dq f_{#1}\left(\bar{\dq{\varXi}}_{1,#1}\right)}%

\global\long\def\b#1#2{\dq b_{#1}\left(\dq{\varXi}_{#1,#2}\right)}%

\global\long\def\adn#1#2#3{\mathrm{Ad}_{#1}\left(#2\right)#3}%

\global\long\def\w#1#2{\underline{\boldsymbol{\mathcal{W}}}_{#1}\left(\dq{\myvec{\Xi}}_{#2}\right)}%

\global\long\def\wadd#1#2#3{\dq{\mathcal{W}}_{#1}\left(\dq{\myvec{\Xi}}_{#2}+\dq{\myvec{\Xi}}_{#3}\right)}%

\global\long\def\fbreve#1{\dq f_{#1}\left(\bar{\dq{\varXi}}_{\breve{1},\breve{#1}}\right)}%

\global\long\def\bbar#1#2{\dq b_{#1}\left(\dq{\varXi}_{\bar{#1},\bar{#2}}\right)}%

\title{Dynamic Modeling of Branched Robots using Modular Composition}
\author{Frederico Fernandes Afonso Silva and Bruno Vilhena Adorno\thanks{F.
F. A. Silva and B. V. Adorno are with the Department of Electrical
and Electronic Engineering and the Manchester Centre for Robotics
and AI, The University of Manchester, Oxford Road, Engineering Building
A, Manchester M13 9PL, United Kingdom (emails: frederico.silva@ieee.org;
bruno.adorno@manchester.ac.uk).}\thanks{This work was partly done
when the authors were affiliated with the Federal University of Minas
Gerais, Belo Horizonte, MG, Brazil, and supported by the Brazilian
agency CAPES, the UK Research and Innovation (UKRI) under the UK government\textquoteright s
Horizon Europe funding guarantee {[}grant number EP/Y024508/1{]},
and the Royal Academy of Engineering under the Research Chairs and
Senior Research Fellowships programme.}}
\maketitle
\begin{abstract}
When modeling complex robot systems such as branched robots, whose
kinematic structures are a tree, current techniques often require
modeling the whole structure from scratch, even when partial models
for the branches are available. This paper proposes a systematic modular
procedure for the dynamic modeling of branched robots comprising several
subsystems, each composed of an arbitrary number of rigid bodies,
providing the final dynamic model by reusing previous models of each
branch. Unlike previous approaches, the proposed strategy is applicable
even if some subsystems are regarded as black boxes, requiring only
twists\textcolor{blue}{{} and their time derivatives,} and wrenches
at the connection points between those subsystems. To help in the
model composition, we also propose a weighted directed graph representation
where the weights encode the propagation of twists\textcolor{blue}{{}
and their time derivatives,} and wrenches between the subsystems.
A simple linear operation on the graph interconnection matrix provides
the dynamics of the whole system. Numerical results using a 24-DoF
fixed-base branched robot composed of eight subsystems show that the
proposed formalism is as accurate as a state-of-the-art library for
robotic dynamic modeling. Additional results using a 30-DoF holonomic
branched mobile manipulator composed of three subsystems demonstrate
the fidelity of our model to a modern robotics simulator and its capability
of dealing with black box subsystems. To further illustrate how the
derived dynamic model can be used in closed-loop control, we also
present a simple formulation of a model-based wrench-driven pose control
for branched robots.
\end{abstract}


\section*{\protect\label{sec:nomenclature}}

\begin{table}
\textcolor{blue}{\noun{Nomenclature}}\textcolor{blue}{}\\

\begin{tabular}{c>{\raggedright}p{0.7\columnwidth}}
\textcolor{blue}{$s$} & \textcolor{blue}{Number of subsystems of the branched robot.}\tabularnewline
\textcolor{blue}{$S$} & \textcolor{blue}{Set containing all subsystems of the branched robot.}\tabularnewline
\textcolor{blue}{$S_{i}\subset S$} & \textcolor{blue}{Set of all subsystems that succeed the $i$th subsystem
of the branched robot.}\tabularnewline
\textcolor{blue}{$m_{i}$} & \textcolor{blue}{Number of elements in the set $S_{i}$.}\tabularnewline
\textcolor{blue}{$i\in S$} & \textcolor{blue}{The $i$th subsystem of the branched robot.}\tabularnewline
\textcolor{blue}{$j\in S_{i}$} & \textcolor{blue}{Subsystem $j$ that is connected to and succeeds
the $i$th subsystem.}\tabularnewline
\textcolor{blue}{$p_{i}\in S$} & \textcolor{blue}{Subsystem $p_{i}$ that is connected to and precedes
the $i$th subsystem.}\tabularnewline
\textcolor{blue}{$a_{i}$} & \textcolor{blue}{Connection point between subsystems $p_{i}$ and
$i$.}\tabularnewline
\textcolor{blue}{$b_{i,j}$} & \textcolor{blue}{Connection point between subsystems $i$ and $j\in S_{i}$.}\tabularnewline
\textcolor{blue}{$n_{i}\in\mathbb{N}$} & \textcolor{blue}{Number of joints/links of the $i$th subsystem.}\tabularnewline
\textcolor{blue}{$\eta\leq n_{i}$} & \textcolor{blue}{Number of joints that precede $b_{i,j}$ in subsystem
$i$.}\tabularnewline
\textcolor{blue}{$n$} & \textcolor{blue}{Number of rigid bodies of the branched robot.}\tabularnewline
\textcolor{blue}{$\mathcal{Q},\dot{\mathcal{Q}},\ddot{\mathcal{Q}}$} & \textcolor{blue}{Sets containing all the joint configurations, velocities,
and accelerations of subsystems that are not black boxes.}\tabularnewline
\textcolor{blue}{$\dq X_{p_{i},i}$} & \textcolor{blue}{Vector containing the rigid transformation between
the connection point $a_{i}$ in the preceding subsystem $p_{i}$
and the CoM of each rigid body within the $i$th subsystem.}\tabularnewline
\textcolor{blue}{$\dq X_{j,i}$} & \textcolor{blue}{Vector containing the rigid transformation between
the connection point }\textbf{\textcolor{blue}{$b_{i,j}$}}\textcolor{blue}{{}
in the succeeding subsystem $j\in S_{i}$ and the $\eta\leq n_{i}$
joints of subsystem $i$.}\tabularnewline
\textcolor{blue}{$\dq{\mathcal{W}}_{i}$} & \textcolor{blue}{Function that maps the stacked vector of total twists
and twist time derivatives at the CoM of each link in the $i$th subsystem
to the corresponding vector of wrenches at the $n_{i}$ joints.}\tabularnewline
\textcolor{blue}{$\dq{\Xi}_{i,i}\in\mathsf{T}^{2n_{i}}$} & \textcolor{blue}{The stacked vector of twists $\overline{\dq{\Xi}}_{i,i}\in\mathsf{T}^{n_{i}}$
and twist time derivatives $\dot{\overline{\dq{\Xi}}}_{i,i}\in\mathsf{T}^{n_{i}}$,
generated by the $i$th subsystem, at the $n_{i}$ CoMs of subsystem
$i\in S$.}\tabularnewline
\textcolor{blue}{$\dq{\Xi}_{p_{i},i}\in\mathsf{T}^{2n_{i}}$} & \textcolor{blue}{The stacked vector of twists $\overline{\dq{\Xi}}_{p_{i},i}\in\mathsf{T}^{n_{i}}$
and twist time derivatives $\dot{\overline{\dq{\Xi}}}_{p_{i},i}\in\mathsf{T}^{n_{i}}$,
expressed in each of the $n_{i}$ CoMs of subsystem $i$, resulting
from the twist at the connection point $a_{i}$.}\tabularnewline
\textcolor{blue}{$\dq{\Xi}_{i}\in\mathsf{T}^{2n_{i}}$} & \textcolor{blue}{The stacked vector of }\textcolor{blue}{\emph{total}}\textcolor{blue}{{}
twists and twists time derivatives at the $n_{i}$ CoMs of subsystem
$i\in S$ (i.e., $\dq{\Xi}_{i}=\dq{\Xi}_{i,i}+\dq{\Xi}_{p_{i},i}$).}\tabularnewline
\textcolor{blue}{$\overline{\dq{\Gamma}}_{j,i}\in\mathsf{W}^{\eta}$} & \textcolor{blue}{The vector of wrenches at the $\eta\leq n_{i}$ joints
of subsystem $i$ due to the wrench at the connection point $b_{i,j}$,
with $j\in S_{i}$.}\tabularnewline
\textcolor{blue}{$\mathring{\dq{\mymatrix{\Gamma}}}_{j,i}\in\mathsf{W}^{n_{i}}$} & \textcolor{blue}{Stacked vector of wrenches, $\mathring{\dq{\mymatrix{\Gamma}}}_{j,i}=\left[\begin{array}{cc}
\overline{\dq{\varGamma}}_{j,i}^{T} & \myvec{\dq 0}_{n_{i}-\eta}^{T}\end{array}\right]^{T}\in\mathsf{W}^{n_{i}}$, that is used to propagate the wrench at the connection point $b_{i,j}$
to all joints of subsystem $i$.}\tabularnewline
\textcolor{blue}{$\dq{\mymatrix{\Gamma}}_{i}\in\mathsf{W}^{n_{i}}$} & \textcolor{blue}{The vector of }\textcolor{blue}{\emph{total}}\textcolor{blue}{{}
wrenches at the $n_{i}$ joints of subsystem $i\in S$.}\tabularnewline
\textcolor{blue}{$\dq{\zeta}_{e_{\ell}}^{\mathcal{L}(\ell)}\in\mathsf{W}$} & \textcolor{blue}{External wrench at the end-effector indexed by $\mathcal{L}(\ell)$
of leaf subsystem $\ell\in S$.}\tabularnewline
\textcolor{blue}{$\dq Z_{e}\in\mathsf{W}^{\ell}$} & \textcolor{blue}{The vector of external wrenches at the $\ell<s$
end-effectors of the $\ell\in S$ leaf subsystems of the branched
robot.}\tabularnewline
\textcolor{blue}{$\dq 0_{n}\in\mathsf{W}^{n}$} & \textcolor{blue}{An $n$-dimensional vector of zeros in the set $\mathsf{W}$.}\tabularnewline
\end{tabular}

\end{table}

\section{Introduction}

In the robotics literature, the Newton-Euler formalism is usually
presented at the level of each rigid body in the mechanical structure,
notably by analyzing the effects of twists and wrenches at the $i$th
link/joint/CoM. This provides a systematic dynamic modeling strategy
applicable to serial manipulators \cite{Siciliano2009,spong2006robot}
and open kinematic trees \cite{Featherstone2008Book,Murray1994,Selig2005}.
However, approaches based on this formalism typically give a monolithic
solution to the system and do not allow model composition. See, for
instance, the formulations presented in \cite{Siciliano2008}.

Monolithic formulations for the dynamic model of branched robots are
not a novelty. Park et al. \cite{Park1995,Park2018a} presented algorithms
based on the Lie algebra associated with the Lie group $\mathrm{SE}(3)$
for the dynamic modeling of open-kinematic chains, whereas Featherstone
\cite{Featherstone1999,Featherstone1999a} proposed divide-and-conquer
algorithms based on spatial algebra, which were later extended by
Mukherjee and Anderson \cite{Mukherjee2007a} to cover flexible bodies.
More recent works on the field of branched robots have focused on
specific dynamic characterizations such as the analysis of a 5-prismatic--spherical--spherical
parallel mechanism \cite{Li2020} and the identification of non-redundant
inertial parameters of branched robots \cite{Tan2022}, contact analysis
\cite{Shafei2018,Ahmadizadeh2021}, motion planning \cite{Wang2018,Glick2022},
and robot design \cite{Liu2023}.

Nonetheless, there are plenty of motivations for seeking a general
formalism for the systematic composition of partial models to obtain
the final model of the complete robotic system. One could assemble
a robot using existing systems whose dynamic models are already known,
such as the limbs of a humanoid robot. In a different scenario, a
self-reconfiguring modular robot \cite{Kotay1998,Neubert2016,Unsal2000}
could possess the dynamic information of its modules that might be
reused. From a control perspective, one could be interested in applying
distributed control strategies to the subsystems comprising a highly
complex dynamic structure composed of thousands of subsystems whose
centralized control would be computationally unfeasible. The few works
in literature focused on modular composition either require the previous
construction of subsystem libraries or demand full knowledge of the
dynamic elements of the whole system.

\subsection{Related works}

The application of linear graph theory in mechanism analysis is not
a novelty either, being used to the dynamic modeling of single rigid
bodies \cite{Andrews1975} and multibody systems composed of open
\cite{Chou1986a} and closed kinematic chains \cite{Sheth1972,Andrews1988,Baciu1990,McPhee1996,McPhee1998,Reungwetwattana2001}.
Most approaches lead to different graphs for the rotational and translational
variables of the mechanism \cite{Andrews1975,Chou1986,Chou1986a,Baciu1990,McPhee1998},
whereas others focus more on computational \cite{Hwang1989} and mechanical
\cite{Althoff2019} aspects than on the dynamic modeling of the system.
Despite exploring several aspects of graph theory, the aforementioned
strategies do not deal with model composition and are, therefore,
monolithic.

To overcome the drawbacks of monolithic approaches, Jain \cite{Jain2012}
proposed a technique of partitioning and aggregating graphs that considers
subsystems in the dynamic modeling of multibody systems. Subgraph
elements are used to compose the mass matrix and the vector of nonlinear
Coriolis and gyroscopic terms of the aggregated system, and then the
stacked vector of generalized forces is calculated. Despite the system
being composed of individual modules, full knowledge of the masses,
inertia tensors, Coriolis accelerations, and gyroscopic terms of the
whole system is required.

McPhee et al. \cite{McPhee2004} presented a strategy that uses individual
subsystem models to derive the dynamic model of mechatronic multibody
systems. Their approach uses free vectors and rotational matrices,
thus decoupling translational and rotational components, which requires
separate graphs. Moreover, that formalism relies on a symbolic implementation,
and each subsystem must be symbolically derived before the modeling
process for the complete system starts.

Moving away from graph representations, Orsino and Hess-Coelho \cite{MatarazzoOrsino2015}
proposed a strategy for the modular modeling of multibody systems,
whose constraints are written as invariants. They use the model of
each subsystem to find the constraint equations among the subsystems.
Then, the constraint equations are used to obtain the system dynamic
equations. Orsino \cite{Orsino2017} extended that formalism by proposing
a hierarchical description of lumped-parameter dynamic systems that
lead to a recursive modeling methodology. Albeit both formulations
\cite{MatarazzoOrsino2015,Orsino2017} do not impose limitations on
how the dynamic equations of each subsystem must be obtained, the
final result is not given in terms of the generalized forces or wrenches
of the mechanism but rather by a system of differential-algebraic
equations of the mechanism's generalized coordinates, constraint equations,
and dynamic equations of motion. Thus, those strategies are not readily
applicable to robotics problems where one typically needs to find
the joint forces/torques as a function of the robot dynamics, which
then are used to design suitable control laws. Moreover, both formulations
\cite{MatarazzoOrsino2015,Orsino2017} consider that the complete
subsystem models are fully available.

More recently, Kumar et al. \cite{Kumar2020} have proposed a modular
solution to the kinematics and dynamics of series-parallel hybrid
robots. Their approach consists of a graph representation in which
edges correspond to rigid bodies and nodes represent the joints connecting
them. The authors use the Lie algebra $se(3)$ to represent twists
and wrenches in the proposed modular recursive Newton-Euler algorithm,
achieving a computationally efficient solution. However, the inverse
dynamics algorithm does not exploit the system's graph representation
with its resultant algebraic operations to find the dynamic model
for the whole system. Furthermore, the current formulation cannot
handle black-box subsystems since twists and wrenches are explicitly
propagated between all subsequent rigid bodies in other subsystems
and must be available for all other modules.

Hess-Coelho et al. \cite{Hess-Coelho2021} presented a dynamic modular
modeling methodology for parallel mechanisms. They use the hierarchical
description proposed by Orsino \cite{Orsino2017} but follow a different
approach to derive the model. Jacobian matrices of the subsystem's
angular and linear velocities are used with the Principle of Virtual
Power to obtain the Euler-Lagrange model of the robot, and modularity
is achieved by using a library of subsystem models. Nonetheless, the
process of obtaining such models is highly dependent on geometric
analysis of the system (i.e., inverse kinematics) and, if no library
containing the subsystems' models is available, the robot dynamics
is found monolithically. Moreover, due to the free-vector representation,
translational and rotational components are decoupled. As Müller \cite{Muller2016b}
pointed out, decoupled representations based on free vectors do not
form a group. Therefore, using free vectors doubles the number of
equations in the problem; also, one needs to explicitly consider lever
arms and their equivalent for velocities and accelerations. In contrast,
in formulations with unified representations, those couplings are
implicitly algebraically found thanks to the machinery of group theory.

Yang et al. \cite{Yang2021} proposed a modular approach for the dynamic
modeling of cable-driven serial robots. They calculate the energy
for each component and then apply an energy-based method to integrate
the components into the complete model of the robotic system. Their
strategy avoids reformulating the coefficient matrix as the number
of modules increases but is not applicable to branched robots containing
black-box subsystems.

In conclusion, the existing model composition strategies oftentimes
generate different graph representations for the translational and
rotational components, increasing the overall complexity, and either
require the previous construction of subsystem libraries or demand
full knowledge of the dynamic elements of the whole system. Some of
them also require a symbolic derivation, as opposed to recursive formulations,
which makes it harder to model reconfigurable systems. This paper
presents a systematic methodology that overcomes those drawbacks.
Table~\ref{tab:summary-differences} presents a summary of the differences
between our proposed formalism and the works in the literature for
the dynamic modeling of branched robots.

\begin{table*}[tph]
\caption{Summary of the differences between our proposed formalism and the
works in literature for the dynamic modeling of branched robots. Check
marks $\left(\checked\right)$ indicate that the topic is covered
in the publication, whereas cross marks $\left(\times\right)$ indicate
it is not.\protect\label{tab:summary-differences}}

\raggedright{}\resizebox{\textwidth}{!}{\footnotesize%
\begin{tabular}{c|c|c|c|c}
\hline 
\textbf{Publication} & \textbf{Graph representation?} & \textbf{Unified graph?} & \textbf{Model composition?} & \textbf{Black-box subsystems?}\tabularnewline
\hline 
\makecell{Park et. al \cite{Park1995,Park2018a}; Featherstone \cite{Featherstone1999,Featherstone1999a};
\\Mukherjee and Anderson \cite{Mukherjee2007a}} & $\times$ & N/A & $\times$ & $\times$\tabularnewline
\hline 
Chou et. al \cite{Chou1986a}; Baciu et. al \cite{Baciu1990}; McPhee
\cite{McPhee1998} & $\checked$ & $\times$ & $\times$ & $\times$\tabularnewline
\hline 
\makecell{Sheth and Uicker \cite{Sheth1972}; Andrews et. al \cite{Andrews1988};
\\Hwang and Haug \cite{Hwang1989}; McPhee \cite{McPhee1996}; \\Reungwetwattana
and Toyama \cite{Reungwetwattana2001}} & $\checked$ & $\checked$ & $\times$ & $\times$\tabularnewline
\hline 
\makecell{Orsino and Hess-Coelho \cite{MatarazzoOrsino2015}; Orsino
\cite{Orsino2017}; \\ Althoff et al. \cite{Althoff2019}; Hess-Coelho
et al. \cite{Hess-Coelho2021}; \\ Yang et al. \cite{Yang2021}} & $\times$ & N/A & $\checked$ & $\times$\tabularnewline
\hline 
Kumar et al. \cite{Kumar2020}; Jain \cite{Jain2012}; McPhee et al.
\cite{McPhee2004} & $\checked$ & $\checked$ & $\checked$ & $\times$\tabularnewline
\hline 
\textbf{\emph{Our proposed formalism}} & $\checked$ & $\checked$ & $\checked$ & $\checked$\tabularnewline
\hline 
\end{tabular}}
\end{table*}

\subsection{Statement of contributions}

This paper presents the following contributions to the state-of-the-art:
\begin{enumerate}
\item A strategy for dynamic model composition, shown in Section~\ref{sec:Model-Composition},
that is applicable even if some subsystems are regarded as black boxes,
requiring only the twists\textcolor{blue}{, twist time derivatives,}
and wrenches at the connection points between different subsystems.
Such information can be obtained either from previous calculations
or sensor readings.
\item A unified graph representation of the system, shown in Section~\ref{subsec:Graph-representation},
that provides the joint wrenches from the calculation of the graph
interconnection matrix, in addition to visually depicting the model
composition.
\item The proposed formulation imposes no restrictions regarding the algebra
used to represent twists and wrenches, as long as some basic properties
are respected. Nonetheless, we present an instantiation for dual quaternion
algebra in Section~\ref{sec:DQ-Model-Composition}.
\item A model-based wrench-driven end-effector motion controller, shown
in Section~\ref{sec:Wrench-Control}, allowing to control all end-effectors
of the branched robot simultaneously. 
\end{enumerate}
We perform numerical evaluations of a fixed-base 24-DOF branched robot
and a 30-DoF holonomic branched mobile manipulator. We then compare
our results with the ones provided by a realistic simulator and Featherstone's
state-of-the-art library for robotic dynamic modeling \cite{Featherstone2008Book}.
To further illustrate how the derived dynamic model can be used in
closed-loop control, we also present a simple formulation of a model-based
wrench control for branched robots.

This paper is organized as follows: Section~\ref{sec:Model-Composition}
presents the proposed dynamic model composition framework; Section~\ref{sec:DQ-Model-Composition}
demonstrates an instantiation of the strategy to dual quaternion algebra;
Section~\ref{sec:Wrench-Control} presents the model-based wrench-driven
end-effector motion controller; Section~\ref{sec:Numerical-evaluation}
shows a numerical evaluation of the proposed methodology, and a comparison
with a state-of-the-art simulator and state-of-the-art library; Section~\ref{sec:Conclusions}
provides the final remarks and points to further research directions;
finally, Appendix~\ref{sec:Appendix-A} briefly reviews the dual
quaternion algebra and Appendix~\ref{sec:Appendix-B} presents the
kinematic and dynamic parameters of the robots used in the simulations.

\section{\protect\label{sec:Model-Composition}Abstract Model Composition}

\begin{figure}
\begin{centering}
\textcolor{white}{\resizebox{1\columnwidth}{!}{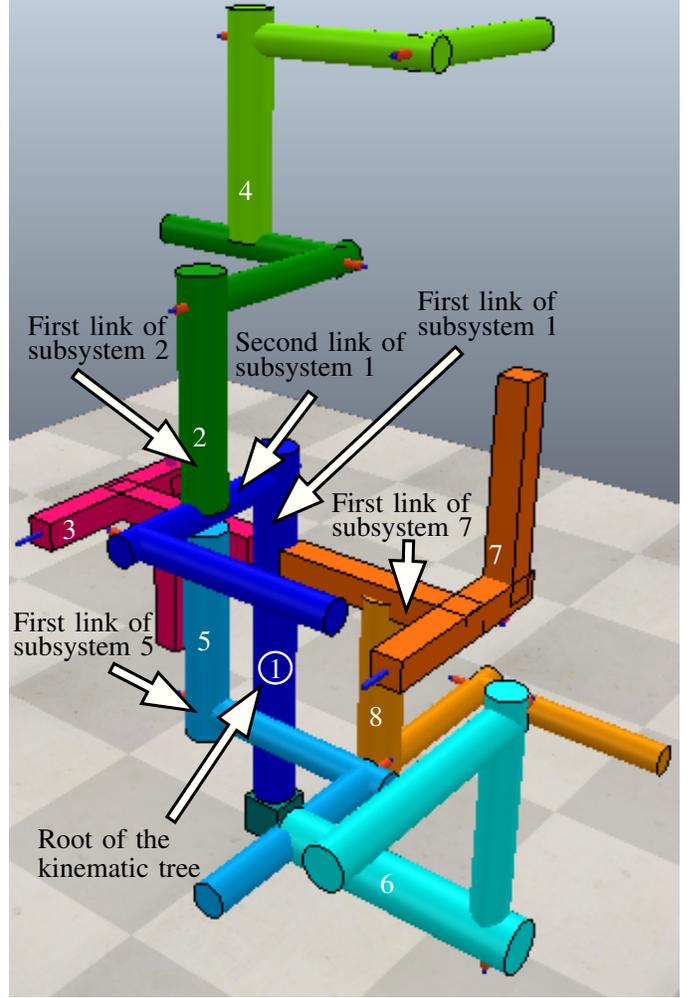}}
\par\end{centering}
\caption{A fixed-base $24$-DoF branched manipulator (BM) composed of eight
subsystems grouped in different colored links associated with indices
from 1 to 8. The root subsystem is indicated by the white circle around
index 1.\protect\label{fig:BM_TRO2023}}
\end{figure}

Consider a branched robot composed of $s$ subsystems\footnote{Subsystems can be defined according to what is convenient for each
problem. For instance, they could be a mobile base, a manipulator,
an off-the-shelf module, or even another branched robot.} shown in Fig.~\ref{fig:BM_TRO2023}. Given the interconnection points
between pair of subsystems and the twists\textcolor{blue}{, twist
time derivatives,} and wrenches applied at those points, our goal
is to obtain a set of equations that describe the whole-body dynamics
as a function of the dynamics of each subsystem $i\in\{1,\ldots,s\}\triangleq S$.

Given the subsystem $i\in S$ that immediately precedes and is connected
to all $j\in\left\{ j_{i,1},\ldots,j_{i,m_{i}}\right\} \triangleq S_{i}\subset S$,
we assume that the first link of subsequent subsystems $j\in S_{i}$
can be connected to any link of the preceding subsystem $i$. For
example, in Fig.~\ref{fig:BM_TRO2023}, the first links of subsystem
3 and 7 are connected to the first link of subsystem 1, whereas the
first links of subsystems 2 and 5 are connected to the second link
of subsystem 1.

Being the complete system an open kinematic tree, each subsystem can
only be preceded by one subsystem but can be succeeded by several
kinematic chains. Therefore, twists \textcolor{blue}{and twist time
derivatives} generated by the subsystem $i$ will be propagated to
each $j\in S_{i}$ and all other subsequent subsystems. On the other
hand, the combined wrenches from all $j\in S_{i}$ will affect $i$
and all its preceding subsystems.

Because the forward propagation of twists \textcolor{blue}{and twist
time derivatives} and the backward propagation of wrenches also happen
within each subsystem, the idea is similar to the classic Newton-Euler
algorithm \cite{Luh1980}. Consider that each subsystem $i\in S$
is composed of $n_{i}$ joints/links, is preceded by a subsystem $p_{i}\in S$,
and is immediately succeeded by all subsystems $j\in S_{i}$. Also,
consider a function 
\begin{equation}
\dq{\mathcal{W}}_{i}\,:\,\mathsf{T}^{2n_{i}}\rightarrow\mathsf{W}^{n_{i}}\label{eq:wrench_function}
\end{equation}
that maps the stacked vector $\dq{\Xi}_{i}\triangleq\left[\begin{array}{cc}
\overline{\dq{\Xi}}_{i}^{T} & \dot{\overline{\dq{\Xi}}}_{i}\end{array}^{T}\right]^{T}\in\mathsf{T}^{2n_{i}}$ of total twists $\overline{\dq{\myvec{\Xi}}}_{i}\in\mathsf{T}^{n_{i}}$
and twist time derivatives $\dot{\overline{\dq{\myvec{\Xi}}}}_{i}\in\mathsf{T}^{n_{i}}$
at the center of mass (CoM) of each link in the $i$th subsystem to
the corresponding vector of total wrenches $\dq{\mymatrix{\Gamma}}_{i}\in\mathsf{W}^{n_{i}}$
at the $n_{i}$ joints.\footnote{The sets $\mathsf{W}$ and $\mathsf{T}$ of wrenches and twists, respectively,
are six-dimensional manifolds that can be represented by different
algebraic structures, such as six-dimensional spatial vectors \cite{Featherstone2008Book},
four-by-four matrices \cite{Murray1994}, or pure dual quaternions
\cite{Adorno2017}.} The wrenches at the joints of each subsystem $i\in S$ originate
from three sources: the twists and their time derivatives at the CoMs
of each link in the $i$th subsystem; the twist and its time derivative
at the connection point $a_{i}$ with the preceding subsystem $p_{i}$
; and the wrenches at the connection points $b_{i,j}$ with each $j\in S_{i}$.
Therefore,
\begin{equation}
\dq{\Gamma}_{i}=\w ii+\sum_{j\in S_{i}}\mathring{\dq{\mymatrix{\Gamma}}}_{j,i},\text{ with }\dq{\Xi}_{i}=\dq{\Xi}_{p_{i},i}+\dq{\Xi}_{i,i},\label{eq:joint_wrenches_subsystem_i}
\end{equation}
where $\dq{\Xi}_{i,i}\triangleq\left[\begin{array}{cc}
\overline{\dq{\Xi}}_{i,i}^{T} & \dot{\overline{\dq{\Xi}}}_{i,i}^{T}\end{array}\right]^{T}\in\mathsf{T}^{2n_{i}}$ is the stacked vector of twists and twist time derivatives at the
$n_{i}$ CoMs of subsystem $i\in S$; the elements of $\dq{\Xi}_{p_{i},i}\triangleq\left[\begin{array}{cc}
\overline{\dq{\Xi}}_{p_{i},i}^{T} & \dot{\overline{\dq{\Xi}}}_{p_{i},i}^{T}\end{array}\right]^{T}\in\mathsf{T}^{2n_{i}}$ are the twists and twist time derivatives at the connection point
$a_{i}$ expressed in each of the $n_{i}$ CoMs of subsystem $i$;
and the elements of $\mathring{\dq{\mymatrix{\Gamma}}}_{j,i}\in\mathsf{W}^{n_{i}}$
are the wrenches of the connection point $b_{i,j}$ with $j\in S_{i}$
expressed in each of the $\eta\leq n_{i}$ joints of subsystem $i$
that precedes $b_{i,j}$. More specifically, $\mathring{\dq{\mymatrix{\Gamma}}}_{j,i}=\left[\begin{array}{cc}
\overline{\dq{\varGamma}}_{j,i}^{T} & \myvec{\dq 0}_{n_{i}-\eta}^{T}\end{array}\right]^{T}\in\mathsf{W}^{n_{i}}$ where $\overline{\dq{\Gamma}}_{j,i}\in\mathsf{W}^{\eta}$ is the
vector of wrenches at the $\eta$ joints preceding the connection
point\footnote{Joints after the connection point are not directly affected by wrenches
from subsequent subsystems. Nonetheless, because those wrenches will
affect the first $\eta$ joints of the $i$th subsystem, the remaining
$n_{i}-\eta$ links will be indirectly affected because the twists
(and their derivatives) at the $\eta$ joints arising from this interaction
will be propagated to the remaining $n_{i}-\eta$ links.} $b_{i,j}$, and $\myvec{\dq 0}_{n_{i}-\eta}\in\mathsf{W}^{n_{i}-\eta}$
is a vector of zeros in the set $\mathsf{W}$. For example, if $\mathsf{W}=\mathbb{R}^{6}$,
then $\myvec{\dq 0}_{n_{i}-\eta}\in\mathsf{W}^{n_{i}-\eta}$ is equivalent
to $\myvec 0_{6(n_{i}-\eta)}\in\mathbb{R}^{6(n_{i}-\eta)}$. If $\mathsf{W}=\mathcal{H}_{p}$,
which is the set of pure dual quaternions \cite{Adorno2017}, then
$\myvec 0_{n_{i}-\eta}\in\mathbb{R}^{n_{i}-\eta}$ because $\mathcal{H}_{p}\ni0\in\mathbb{R}$
as $\mathbb{R}\subset\mathcal{H}_{p}$.

\begin{figure}
\subfloat[Forward propagation of twists from subsystem $p_{i}$ to the remaining
subsystems of the branched robot. Hatched regions indicate the areas
influenced by the twist at the connection point $a_{i}$.]{\begin{centering}
\resizebox{0.95\columnwidth}{!}{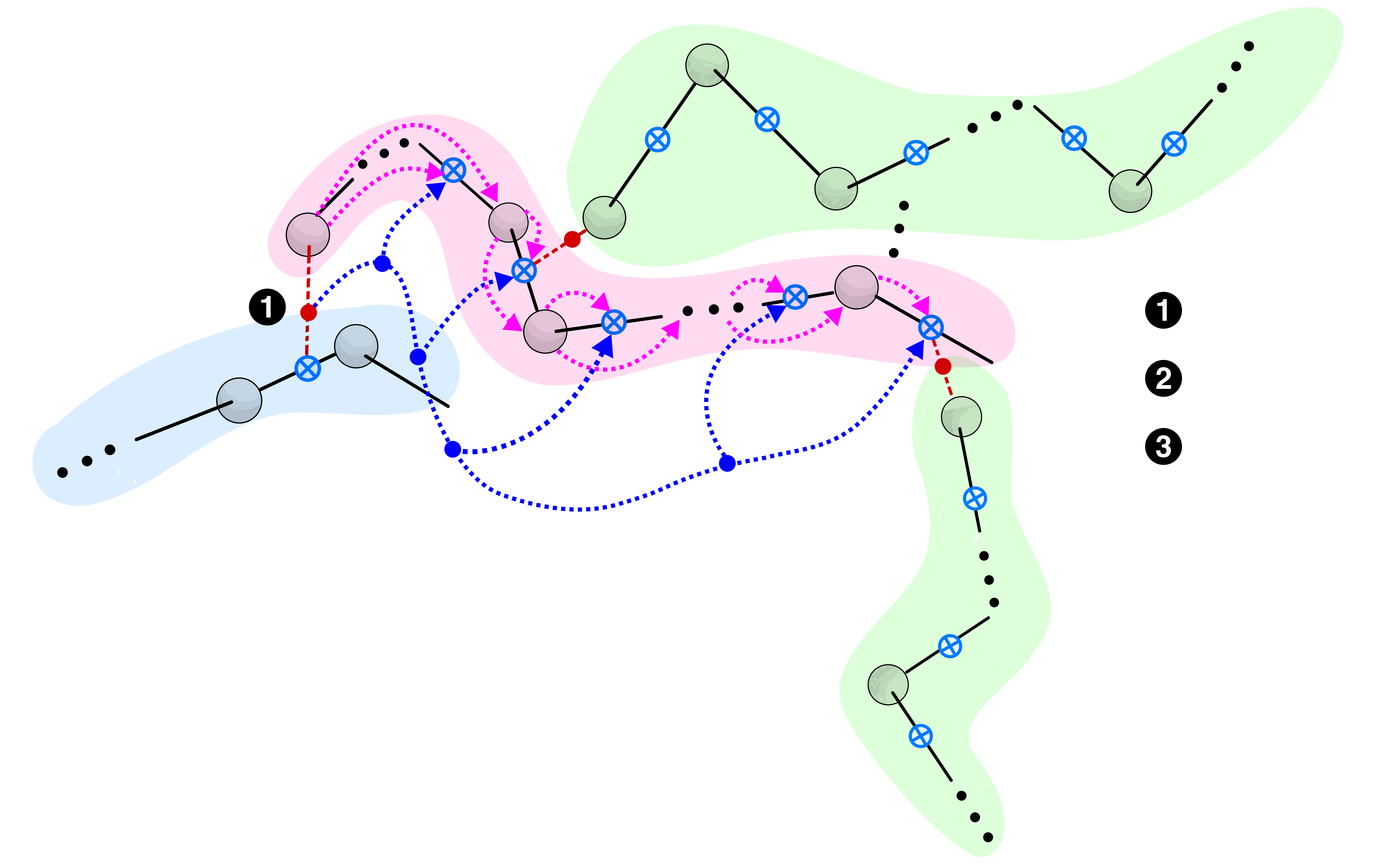}
\par\end{centering}
}\quad{}\subfloat[Backward propagation of wrenches from subsystems $j\in S_{i}$ to
the remaining subsystems of the branched robot. Hatched and crosshatched
regions indicate the areas influenced by the wrenches at the connection
points $b_{i,j}$.]{\begin{centering}
\resizebox{0.95\columnwidth}{!}{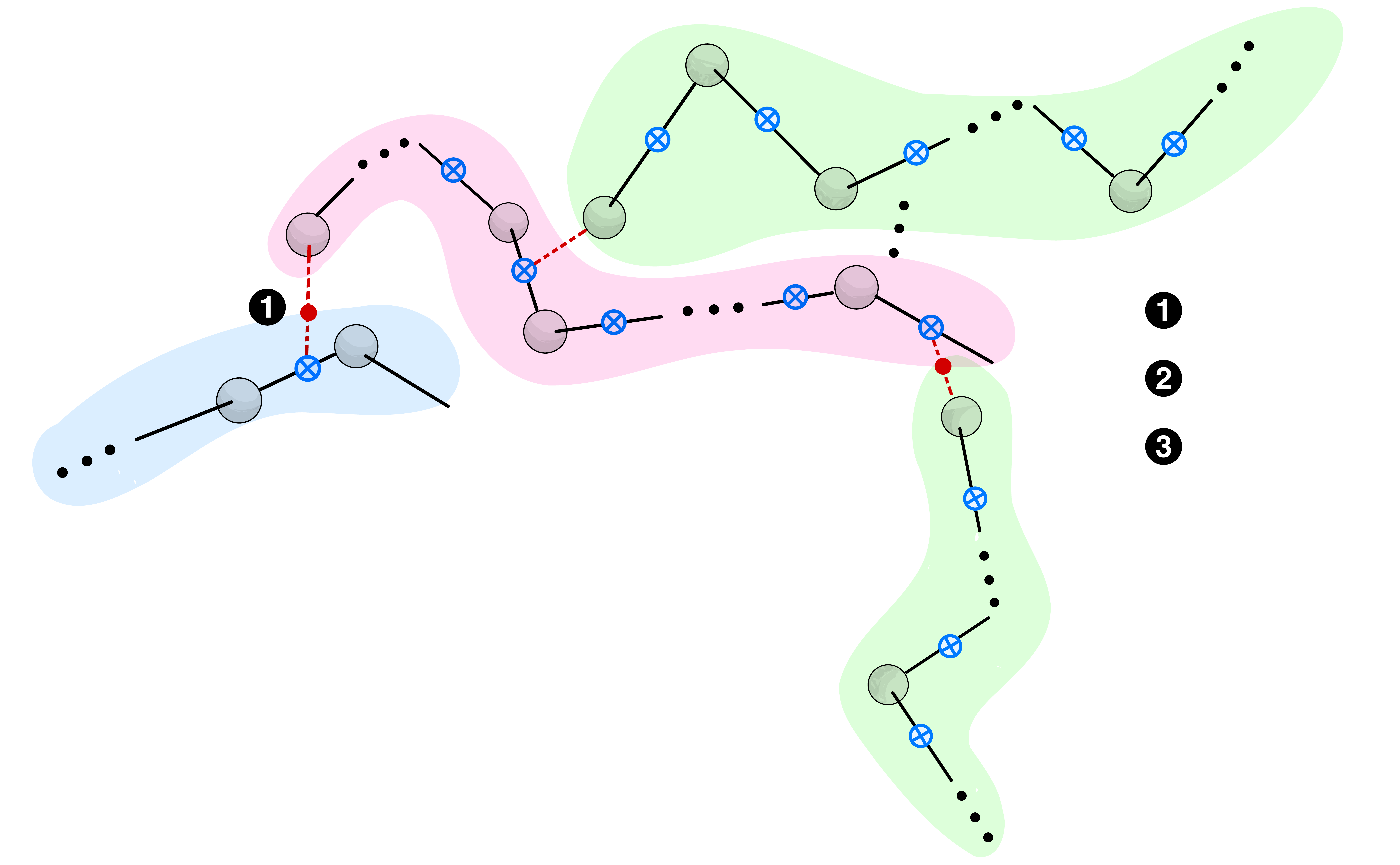}
\par\end{centering}
}

{\footnotesize\caption{Wrenches generated at the joints of the $i$th subsystem (pink region).
For each subsystem, the large gray circles represent joints, solid
black lines represent links, and blue crossed circles represent CoMs.
The preceding subsystem $p_{i}$ is given in blue, the subsequent
subsystems $j\in S_{i}$ are colored in green, and red circles on
the red dashed lines, numbered from 1 to 3, indicate the connection
points. Dotted arrows represent the forward propagation of twists,
whereas solid arrows represent the backward propagation of wrenches.\protect\label{fig:Wrenches-generated-at-the-ith-subsystem}}
}{\footnotesize\par}
\end{figure}

The wrenches generated at the joints of the $i$th subsystem as a
result of its own motion, the motion of its predecessor, and the wrenches
from its successors is shown in Fig.~\ref{fig:Wrenches-generated-at-the-ith-subsystem}.
The connection point $a_{i}$ between $p_{i}$ and the $i$th subsystem,
denoted by \circled{1} in Fig.~\ref{fig:Wrenches-generated-at-the-ith-subsystem},
contains the \emph{resultant} twist generated by all moving joints
from the connection point up to the root node. On the other hand,
the connection points $b_{i,j_{i,1}}$ and $b_{i,j_{i,2}}$ between
subsequent kinematic chains connected to the $i$th subsystem, respectively
denoted by \circled{2} and \circled{3} in Fig.~\ref{fig:Wrenches-generated-at-the-ith-subsystem},
contain the \emph{resultant} wrenches generated by those systems and
all their successors.

Therefore, to calculate the wrenches at the joints of the $i$th subsystem,
we need the information of the twists and their time derivatives at
its CoMs, the twist and its time derivative at the connection point
with its predecessor, and the wrenches at the connection points with
its successors. Thus, even when some subsystems are presented as black
boxes, the dynamics of the overall system can still be obtained, as
long as we have the information of twists, twists derivatives, and
wrenches at the connection points (e.g., through sensor readings).\textcolor{blue}{{}
}This property is particularly relevant in modular robotics applications.
For instance, suppose a new, unknown module is appended to a self-reconfiguring
robot. The proposed formalism would allow the reconfigurable robot
to update its complete dynamic model despite the newly appended black
box subsystem as long as wrenches and twists are measured at the connection
points. Conversely, methods that require complete knowledge of the
newly appended module dynamics would be unfit to handle this situation.

\begin{example}
\label{exa:two_subsystems}For the sake of simplicity, let us consider
only subsystems 1 and 2 in Fig.~\ref{fig:BM_TRO2023} and disregard
the remaining subsystems. In that case, the vectors $\dq{\Gamma}_{1}\in\mathsf{W}^{n_{1}}$
and $\dq{\Gamma}_{2}\in\mathsf{W}^{n_{2}}$ of wrenches at the joints
of subsystems 1 and 2, respectively, with $n_{1}=n_{2}=3$, are given
by
\begin{align}
\dq{\Gamma}_{1} & =\w 11+\mathring{\dq{\mymatrix{\Gamma}}}_{2,1},\nonumber \\
\dq{\Gamma}_{2} & =\w 22,\label{eq:joint_wrenches_subsystems_1_and_2}
\end{align}
where $\dq{\Xi}_{1}=\dq{\Xi}_{1,1}$ and $\dq{\Xi}_{2}=\dq{\Xi}_{1,2}+\dq{\Xi}_{2,2}$.

Since subsystem 1 is the first in the chain, the wrenches at its joints
are generated by the twists and their time derivatives at the CoMs
of its own links, and by the reaction wrenches generated by subsystem
2. On the other hand, for subsystem 2, which is the last in the chain
if we disregard the other subsystems, the wrenches at the joints are
generated by the twists and their time derivatives at the CoMs of
its own links, in addition to the the twist and twist derivative that
subsystem 1 contributes through the connection point.
\end{example}

\subsection{Graph representation\protect\label{subsec:Graph-representation}}

\begin{figure*}
\begin{centering}
\setlength{\tabcolsep}{1pt}%
\begin{tabular}{>{\centering}m{0.2\textwidth}>{\centering}m{0.06\textwidth}>{\centering}m{0.2\textwidth}>{\centering}m{0.06\textwidth}>{\centering}m{0.4\textwidth}}
\begin{centering}
\resizebox{!}{0.19\textheight}{
\begingroup%
  \makeatletter%
  \providecommand\color[2][]{%
    \errmessage{(Inkscape) Color is used for the text in Inkscape, but the package 'color.sty' is not loaded}%
    \renewcommand\color[2][]{}%
  }%
  \providecommand\transparent[1]{%
    \errmessage{(Inkscape) Transparency is used (non-zero) for the text in Inkscape, but the package 'transparent.sty' is not loaded}%
    \renewcommand\transparent[1]{}%
  }%
  \providecommand\rotatebox[2]{#2}%
  \newcommand*\fsize{\dimexpr\f@size pt\relax}%
  \newcommand*\lineheight[1]{\fontsize{\fsize}{#1\fsize}\selectfont}%
  \ifx\svgwidth\undefined%
    \setlength{\unitlength}{102.20541045bp}%
    \ifx\svgscale\undefined%
      \relax%
    \else%
      \setlength{\unitlength}{\unitlength * \real{\svgscale}}%
    \fi%
  \else%
    \setlength{\unitlength}{\svgwidth}%
  \fi%
  \global\let\svgwidth\undefined%
  \global\let\svgscale\undefined%
  \makeatother%
  \begin{picture}(1,1.2517125)%
    \lineheight{1}%
    \setlength\tabcolsep{0pt}%
    \put(0,0){\includegraphics[width=\unitlength,page=1]{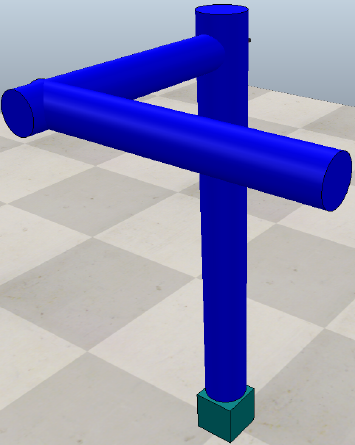}}%
    \put(0.60305993,0.32831156){\color[rgb]{1,1,1}\makebox(0,0)[lt]{\lineheight{1.25}\smash{\begin{tabular}[t]{l}1\end{tabular}}}}%
  \end{picture}%
\endgroup%
}
\par\end{centering}
\smallskip{}

 & %
\begin{minipage}[t]{0.05\textwidth}%
\begin{center}
\resizebox{1\textwidth}{!}{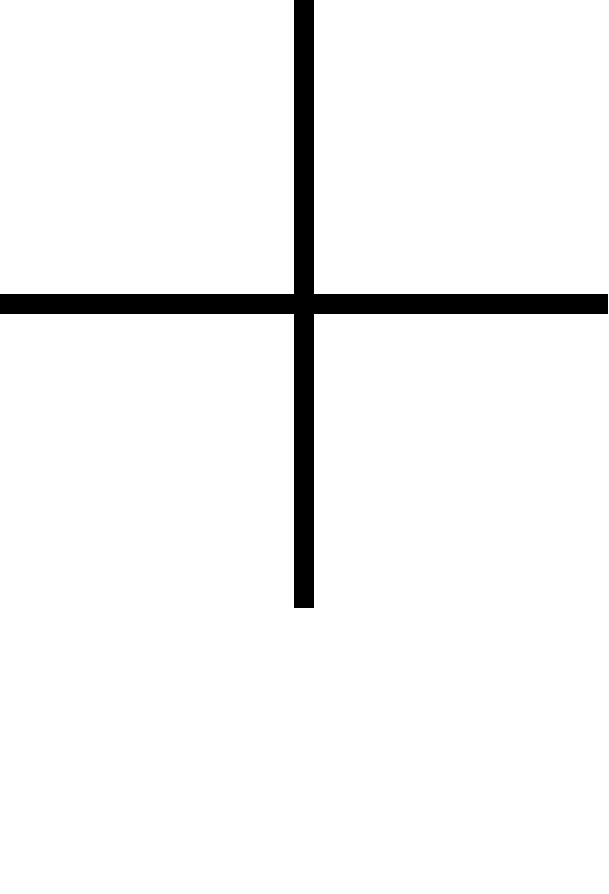}
\par\end{center}%
\end{minipage} & \begin{centering}
\resizebox{!}{0.2\textheight}{
\begingroup%
  \makeatletter%
  \providecommand\color[2][]{%
    \errmessage{(Inkscape) Color is used for the text in Inkscape, but the package 'color.sty' is not loaded}%
    \renewcommand\color[2][]{}%
  }%
  \providecommand\transparent[1]{%
    \errmessage{(Inkscape) Transparency is used (non-zero) for the text in Inkscape, but the package 'transparent.sty' is not loaded}%
    \renewcommand\transparent[1]{}%
  }%
  \providecommand\rotatebox[2]{#2}%
  \newcommand*\fsize{\dimexpr\f@size pt\relax}%
  \newcommand*\lineheight[1]{\fontsize{\fsize}{#1\fsize}\selectfont}%
  \ifx\svgwidth\undefined%
    \setlength{\unitlength}{76.80119732bp}%
    \ifx\svgscale\undefined%
      \relax%
    \else%
      \setlength{\unitlength}{\unitlength * \real{\svgscale}}%
    \fi%
  \else%
    \setlength{\unitlength}{\svgwidth}%
  \fi%
  \global\let\svgwidth\undefined%
  \global\let\svgscale\undefined%
  \makeatother%
  \begin{picture}(1,1.49893836)%
    \lineheight{1}%
    \setlength\tabcolsep{0pt}%
    \put(0,0){\includegraphics[width=\unitlength,page=1]{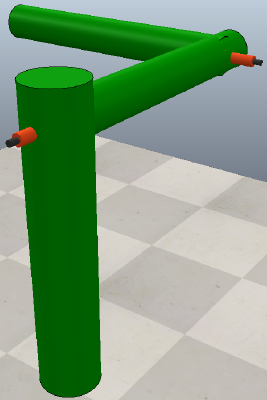}}%
    \put(0.21469442,0.16020167){\color[rgb]{1,1,1}\makebox(0,0)[lt]{\lineheight{1.25}\smash{\begin{tabular}[t]{l}2\end{tabular}}}}%
  \end{picture}%
\endgroup%
}
\par\end{centering}
\smallskip{}

 & %
\begin{minipage}[t]{0.05\textwidth}%
\begin{center}
\resizebox{1\textwidth}{!}{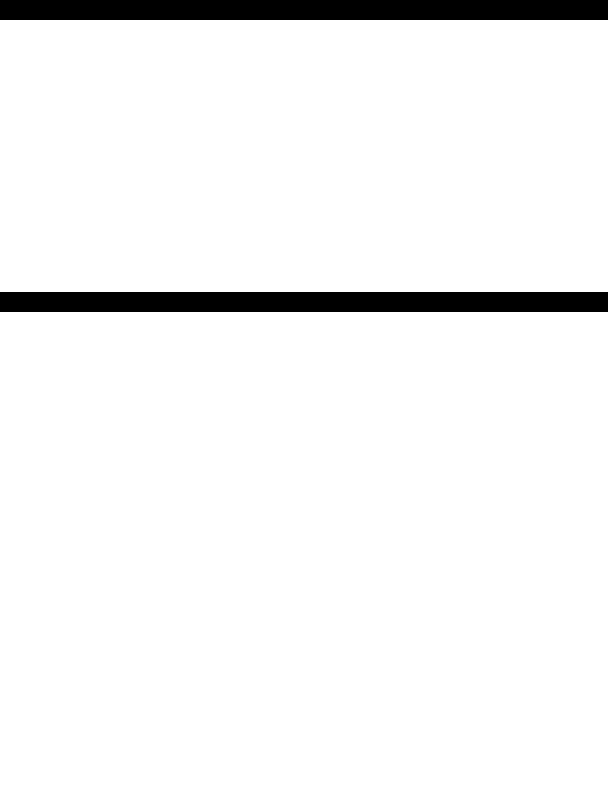}
\par\end{center}%
\end{minipage} & \resizebox{!}{0.2\textheight}{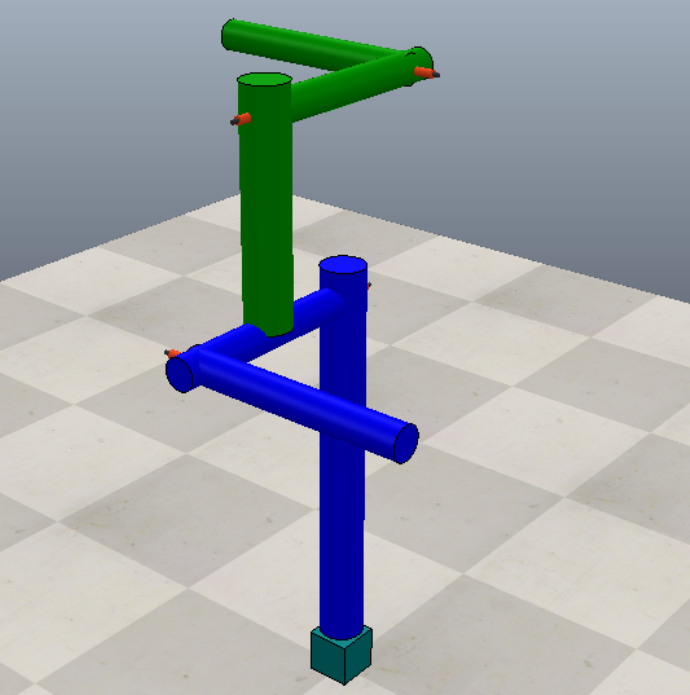}

\smallskip{}

\tabularnewline
\end{tabular}
\par\end{centering}
\caption{Assembly process of a robot composed of different subsystems: (a)
$3$-DoF robot given by the first subsystem shown in Fig.~\ref{fig:BM_TRO2023}
and the corresponding graph representation---the wrenches $\protect\w 11$
of the system are exclusively generated by its own twists \textcolor{blue}{and
twist time derivatives}; (b) $3$-DoF robot given by the second subsystem
shown in Fig.~\ref{fig:BM_TRO2023} and its graph representation---the
wrenches $\underline{\boldsymbol{\mathcal{W}}}_{2}(\protect\dq{\Xi}_{2,2})$
of the system are exclusively generated by its own twists \textcolor{blue}{and
twist derivatives}; (c) $6$-DoF assembled robot given by the first
two subsystems shown in Fig.~\ref{fig:BM_TRO2023} and its graph
representation. In this combined system, the wrenches of system 2
originates from its twists\textcolor{blue}{{} and twist derivatives}
$\protect\dq{\Xi}_{2,2}$ and the twist\textcolor{blue}{{} and twist
derivatives} $\protect\dq{\Xi}_{1,2}$ at the connection with system
1. Similarly, the wrenches at the joints of system 1 result from the
self-motion wrenches $\protect\w 11$ of system 1 added by the wrench
$\mathring{\protect\dq{\protect\mymatrix{\Gamma}}}_{2,1}$ at its
connection with system 2.\protect\label{fig:graph_subsystems_i_and_ji}}
\end{figure*}

Each subsystem in a branched robot may be represented as a vertex
in a graph, in which directed, weighted edges represent the propagation
of wrenches, twists, and twist time derivatives. The advantage of
such representation is that in addition to visually depicting the
model composition, it provides the joint wrenches from the calculation
of the graph wrench interconnection matrix. For instance, the weighted
graph in Fig.~\ref{fig:graph_subsystems_i_and_ji} represents the
system of Example~\ref{exa:two_subsystems}, where dashed edges correspond
to the propagation of twists and their time derivatives, and solid
edges represent the propagation of wrenches.

The wrench interconnection matrix of the graph presented in Fig.~\ref{fig:graph_subsystems_i_and_ji}
is given by\footnote{The interconnection matrix is always $\dq{\mymatrix A}\in\mathsf{W}^{n\times s}$,
where $n$ is the total number of rigid bodies in the system and $s$
is the total number of subsystems of the branched robot.}
\begin{equation}
\dq{\mymatrix A}=\left[\dq A_{ij}\right]\triangleq\left[\begin{array}{cc}
\w 11 & \mathring{\dq{\mymatrix{\Gamma}}}_{2,1}\\
\dq{\myvec 0}_{3} & \w 22
\end{array}\right]\in\mathsf{W}^{6\times2},\label{eq:adjacency_matrix_ni+nj-DOF_robot}
\end{equation}
where $\dq{\Xi}_{1}=\dq{\Xi}_{1,1}$ and $\dq{\Xi}_{2}=\dq{\Xi}_{1,2}+\dq{\Xi}_{2,2}$.
The vectors $\dq A_{ij}\in\mathsf{W}^{n_{i}}$ of the partitioned
matrix $\dq{\mymatrix A}$ indicate the wrench propagation from vertex
$j$ to vertex $i$, represented as a solid edge. Because there is
no wrench propagation from subsystem 1 to 2 in Example~\ref{exa:two_subsystems},
$\dq A_{2,1}=\dq{\myvec 0}_{3}\in\mathsf{W}^{3}$, which is a vector
of zeros in the set $\mathsf{W}$. Therefore, since the corresponding
``weight'' is zero, there is no solid edge from 1 to 2 in the graph
on Fig.~\ref{fig:graph_subsystems_i_and_ji}. The wrench interconnection
matrix $\dq{\mymatrix A}$ is analogous to the weighted adjacency
matrix of the graph, but the element $\dq A_{ij}$ provides the joint
wrenches imposed by one subsystem onto another instead of a real scalar
weight.

\begin{figure*}
\begin{centering}
\resizebox{0.8\paperwidth}{!}{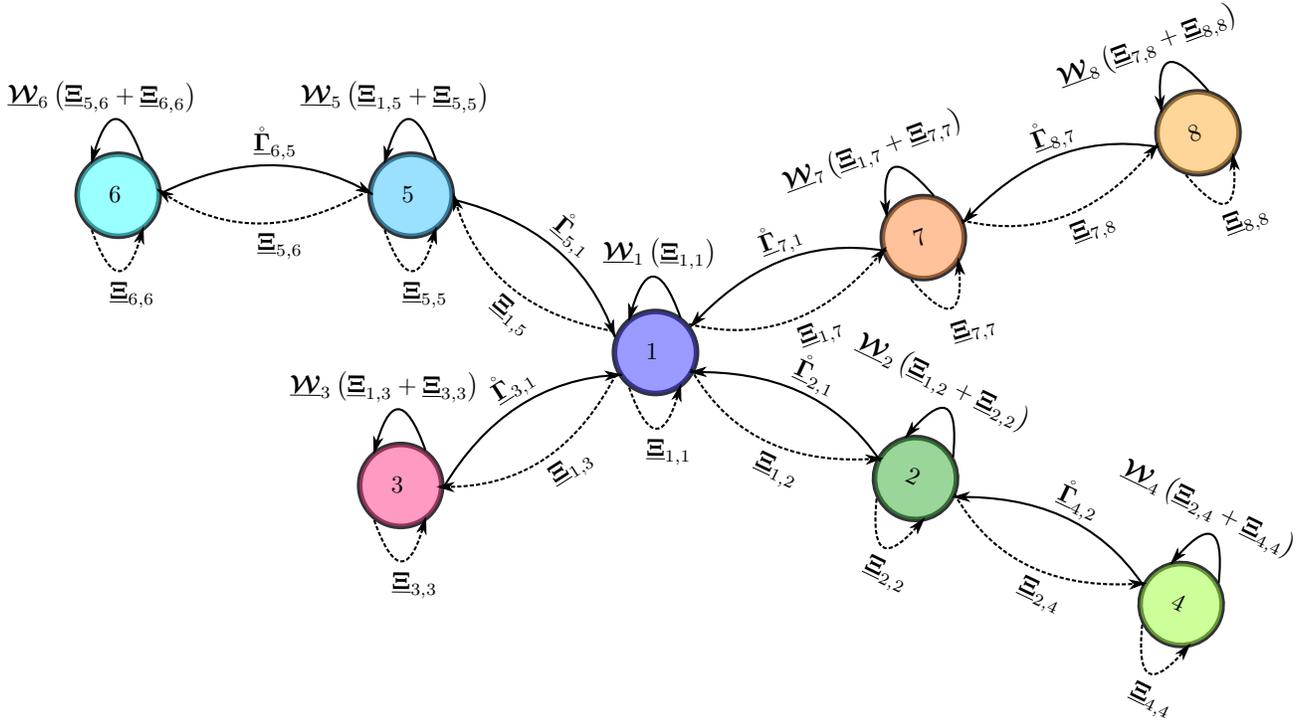}
\par\end{centering}
\caption{Graph representation of the $24$-DoF branched robot. The colored
nodes follow the color scheme adopted in Fig.~\ref{fig:BM_TRO2023}.
Only the root subsystem, represented by vertex 1, has just one incoming
dashed edge, meaning that no twists \textcolor{blue}{and twist time
derivatives} are propagated from other subsystems. On the other hand,
only the vertices corresponding to the leaves of the branched system
(3, 4, 6, and 8) have only one incoming solid edge, meaning that no
wrenches are propagated from other subsystems.\protect\label{fig:BM_TRO2023_graph}}
\end{figure*}

More generally, the graph representation of the complete system composed
of $s$ kinematic chains is constructed as follows:
\begin{enumerate}
\item Create a vertex for each kinematic chain.
\item Add the edges, according to the following rules:
\begin{enumerate}
\item Each vertex $i$ has a dashed edge self-loop weighted by its own stacked
vector of twists and twist time derivatives $\dq{\Xi}_{i,i}\in\mathsf{T}^{2n_{i}}$;
\item Except for the vertex representing the root subsystem of the branched
kinematic chain, each vertex $i$ has an incoming dashed edge from
the vertex $p_{i}$ representing its predecessor, ``weighted'' by
the stacked vector of twists and twist time derivatives $\dq{\Xi}_{p_{i},i}\in\mathsf{T}^{2n_{i}}$,
and a solid edge self-loop weighted by its own vector of wrenches
given by $\dq{\mathcal{W}}_{i}(\dq{\Xi}_{i,i}+\dq{\Xi}_{p_{i},i})\in\mathsf{W}^{n_{i}}$;
\begin{enumerate}
\item If the $i$th vertex represents the root subsystem of the branched
kinematic chain, then the solid edge self-loop is weighted by $\dq{\mathcal{W}}_{i}(\dq{\Xi}_{i,i})\in\mathsf{W}^{n_{i}}$.
\end{enumerate}
\item Except for the vertex representing the root subsystem of the branched
kinematic chain, each vertex $i$ has an outgoing solid edge that
goes to the vertex $p_{i}$ representing its predecessor, ``weighted''
by the vector of wrenches $\mathring{\dq{\mymatrix{\Gamma}}}_{i,p_{i}}\in\mathsf{W}^{n_{p_{i}}}$.
\end{enumerate}
\end{enumerate}
\textcolor{blue}{}
\begin{example}
\label{exa:BM_TRO2023_wrenches_from_A}Consider the $24$-DoF branched
robot shown in Fig.~\ref{fig:BM_TRO2023}, where $n_{1}=n_{2}=\cdots=n_{8}=3$.
Following the procedure described previously, the weighted graph representing
this robot is given in Fig.~\ref{fig:BM_TRO2023_graph}. Moreover,
the interconnection matrix $\dq{\mymatrix A}\in\mathsf{W}^{24\times8}$
is given by
\begin{multline*}
\dq A=\\
{\scriptstyle \left[\begin{array}{cccccccc}
\!\dq{\mathcal{W}}_{1} & \!\mathring{\dq{\mymatrix{\Gamma}}}_{2,1} & \!\mathring{\dq{\mymatrix{\Gamma}}}_{3,1} & \!\dq{\myvec 0}_{3} & \!\mathring{\dq{\mymatrix{\Gamma}}}_{5,1} & \!\dq{\myvec 0}_{3} & \!\mathring{\dq{\mymatrix{\Gamma}}}_{7,1} & \!\dq{\myvec 0}_{3}\\
\!\dq{\myvec 0}_{3} & \!\dq{\mathcal{W}}_{2} & \!\dq{\myvec 0}_{3} & \!\mathring{\dq{\mymatrix{\Gamma}}}_{4,2} & \!\dq{\myvec 0}_{3} & \!\dq{\myvec 0}_{3} & \!\dq{\myvec 0}_{3} & \!\dq{\myvec 0}_{3}\\
\!\dq{\myvec 0}_{3} & \!\dq{\myvec 0}_{3} & \!\dq{\mathcal{W}}_{3} & \!\dq{\myvec 0}_{3} & \!\dq{\myvec 0}_{3} & \!\dq{\myvec 0}_{3} & \!\dq{\myvec 0}_{3} & \!\dq{\myvec 0}_{3}\\
\!\dq{\myvec 0}_{3} & \!\dq{\myvec 0}_{3} & \!\dq{\myvec 0}_{3} & \!\dq{\mathcal{W}}_{4} & \!\dq{\myvec 0}_{3} & \!\dq{\myvec 0}_{3} & \!\dq{\myvec 0}_{3} & \!\dq{\myvec 0}_{3}\\
\!\dq{\myvec 0}_{3} & \!\dq{\myvec 0}_{3} & \!\dq{\myvec 0}_{3} & \!\dq{\myvec 0}_{3} & \!\dq{\mathcal{W}}_{5} & \!\mathring{\dq{\mymatrix{\Gamma}}}_{6,5} & \!\dq{\myvec 0}_{3} & \!\dq{\myvec 0}_{3}\\
\!\dq{\myvec 0}_{3} & \!\dq{\myvec 0}_{3} & \!\dq{\myvec 0}_{3} & \!\dq{\myvec 0}_{3} & \!\dq{\myvec 0}_{3} & \!\dq{\mathcal{W}}_{6} & \!\dq{\myvec 0}_{3} & \!\dq{\myvec 0}_{3}\\
\!\dq{\myvec 0}_{3} & \!\dq{\myvec 0}_{3} & \!\dq{\myvec 0}_{3} & \!\dq{\myvec 0}_{3} & \!\dq{\myvec 0}_{3} & \!\dq{\myvec 0}_{3} & \!\dq{\mathcal{W}}_{7} & \!\mathring{\dq{\mymatrix{\Gamma}}}_{8,7}\\
\!\dq{\myvec 0}_{3} & \!\dq{\myvec 0}_{3} & \!\dq{\myvec 0}_{3} & \!\dq{\myvec 0}_{3} & \!\dq{\myvec 0}_{3} & \!\dq{\myvec 0}_{3} & \!\dq{\myvec 0}_{3} & \!\dq{\mathcal{W}}_{8}
\end{array}\right]},
\end{multline*}
in which
\[
\dq{\mathcal{W}}_{i}=\begin{cases}
\w i{i,i} & \text{if }i=1,\\
\wadd i{p_{i},i}{i,i} & \text{if }i\in\{2,\ldots,8\},
\end{cases}
\]
$p_{2}=p_{3}=p_{5}=p_{7}=1$, $p_{4}=2$, $p_{6}=5$, and $p_{8}=7$,
and $\dq{\myvec 0}_{3}\in\mathsf{W}^{3}$ is a vector of zeros in
$\mathsf{W}^{3}$.
\end{example}

The proposition below shows how the adjacency matrix is used to derive
the model of the complete assembled system.
\begin{prop}
\label{prop:assembled_model_from_adjacency_matrix}Let a branched
kinematic system be composed of $n$ rigid bodies divided into a set
of $s$ coupled subsystems, each one containing $n_{1}$, $n_{2}$,
$\ldots$, $n_{s}$ rigid bodies, respectively. Considering the proposed
weighted graph representation with its corresponding adjacency matrix
(\ref{eq:adjacency_matrix_ni+nj-DOF_robot}), the vector of joint
wrenches $\dq{\mymatrix{\Gamma}}_{\mathrm{total}}$ of the complete
system is given by
\begin{equation}
\dq{\mymatrix{\Gamma}}_{\mathrm{total}}=\left[\begin{array}{cccc}
\dq{\Gamma}_{1}^{T} & \dq{\Gamma}_{2}^{T} & \cdots & \dq{\Gamma}_{s}^{T}\end{array}\right]^{T}=\dq{\mymatrix A}\dq{\myvec 1}_{s}\in\mathsf{W}^{n},\label{eq:wrench_t_through_Ax1}
\end{equation}
where $\dq{\Gamma}_{i}\in\mathsf{W}^{n_{i}}$ is the vector of the
total joint wrenches of the $i$th subsystem, $\dq{\mymatrix A}\in\mathsf{W}^{n\times s}$
is the interconnection matrix, with $n=\sum_{i=1}^{s}n_{i}$, and
$\dq{\myvec 1}_{s}\in\mathsf{W}^{s}$ is an $s$-dimensional vector
of identity elements (under the multiplication operation) in the set
$\mathsf{W}$.
\end{prop}
\begin{IEEEproof}
Each block element $\dq{\myvec A}_{ij}\in\mathsf{W}^{n_{j}}$ of the
matrix $\dq{\mymatrix A}$ represents the weight of the edge from
vertex $j$ to vertex $i$ in the interconnection graph and, therefore,
the propagation of wrenches from subsystem $j$ to $i$. Hence, each
row of $\dq{\mymatrix A}$ contains all the wrenches acting upon the
joints of the $i$th subsystem. Therefore, the vector $\dq{\Gamma}_{i}$
of the total joint wrenches of the $i$th subsystem is given by $\dq{\Gamma}_{i}=\sum_{j=1}^{s}\dq{\myvec A}_{ij}=\sum_{j=1}^{s}\left(\dq{\myvec A}_{ij}\cdot\myvec 1\right)$,
where $\myvec 1\in\mathsf{W}$ is the identity element in $\mathsf{W}$
under the multiplication operation such that $\dq{\myvec A}_{ij}\cdot\myvec 1=\dq{\myvec A}_{ij}$
. Thus, $\dq{\mymatrix{\Gamma}}_{\mathrm{total}}=\dq{\mymatrix A}\dq{\myvec 1}_{s}$.\footnote{If $\mathsf{W}=\mathbb{R}^{6}$ or $\mathsf{W}=\mathcal{H}_{p}$ (i.e.,
a pure dual quaternion), then $\myvec 1=1$ and hence $\dq{\myvec 1}_{s}=\myvec 1_{s}\in\mathbb{R}^{s}$.
If $\mathsf{W}\in\mathrm{se}^{*}(3)$, then $\myvec 1=\mymatrix I_{4\times4}$
and hence $\dq{\myvec 1}_{s}\in\mathbb{R}^{4s\times4}$.}
\end{IEEEproof}
Algorithms \ref{alg:DMC}, \ref{alg:DMC_FR}, and \ref{alg:N} summarize
the proposed modular dynamic modeling formalism. The inputs for the
dynamic modular composition (DMC) algorithm are the sets $\mathcal{Q}$,
$\dot{\mathcal{Q}}$, and $\ddot{Q}$, respectively, comprised of
the joint configurations, velocities, and accelerations of subsystems
that are not black boxes. Algorithm~\ref{alg:DMC_FR} presents the
forward propagation of the stacked vector of twists and twist time
derivatives $\dq{\Xi}\in\mathsf{T}^{2n}$ at the $n=\sum_{i=1}^{s}n_{i}$
CoMs of the robot, in which a breadth-first search (BFS) algorithm
is used to traverse the tree \cite{LaValle2006book}. Since the graph
representing the topology of the branched robot is constant, we assume
it is internally available. Between lines \ref{line:begin_Xi_=00007Bi,i=00007D}
and \ref{line:end_Xi_=00007Bi,i=00007D}, the twists $\dq{\Xi}_{i,i}$
are calculated due to the subsystem's own motion. The implementation
of \noun{forward\_recursion} in line~\ref{line:forward_recursion}
depends on the algebra used to represent wrenches and twists. For
instance, we provide a formulation using dual quaternion algebra in
Algorithm~\ref{alg:dqNE_forward_recursion}. In line \ref{line:calculate_X_pj_j},
the poses \textcolor{blue}{$\dq X_{p_{j},j}$} associated with each
of the subsystem's CoMs are calculated as a function of the subsystem's
configuration \textcolor{blue}{$\myvec q_{j}\in\mathcal{Q}$}. Between
lines \ref{line:begin_Xi_=00007Bpi,i=00007D} and \ref{line:end_Xi_=00007Bpi,i=00007D},
the twists $\dq{\Xi}_{p_{j},j}$ due to the preceding subsystem's
motion is obtained. If subsystem $p_{j}$ is a black box, the twist
$\dq{\xi}_{0,a_{j}}^{a_{j}}$ at the connection point $a_{j}$ is
obtained from sensor readings. Alternatively, the values calculated
internally in subsystem $p_{j}$ are used to calculate $\dq{\xi}_{0,a_{j}}^{a_{j}}$.\footnote{\textcolor{blue}{If the configuration of subsystem $p_{j}$ has not
changed, previously calculated and stored values can be used instead
of recalculating them.}} The twist $\dq{\xi}_{0,a_{j}}^{a_{j}}$ is then projected to the
appropriate CoM references frames of subsystem $p_{j}$. This operation
is also algebra dependent. For dual quaternions, this transformation
is given by (\ref{eq:vector_twists_p}) and (\ref{eq:vector_twists_dot_p}).

Algorithm~\ref{alg:N} calculates the vector $\dq{\myvec{\Gamma}}\in\mathsf{W}^{n}$
containing the wrenches at the $n$ joints of the branched robot.
For that, the non-zero elements of the interconnection matrix $\dq A$,
given as in (\ref{eq:adjacency_matrix_ni+nj-DOF_robot}), are calculated
from right to left. Those are summed to yield each $\dq{\myvec{\Gamma}}_{i}\subseteq\dq{\myvec{\Gamma}}$
(i.e., $\dq{\myvec{\Gamma}}_{i}$ is the sum of the coefficients of
the $i$th row of $\dq A$). Between lines \ref{line:begin_Wi} and
\ref{line:end_Wi}, algorithm $\dq{\mathcal{N}}$ obtains the wrenches
of subsystem $i$ due to its own motion, whereas between lines \ref{line:begin_Gamma_=00007Bj,i=00007D}
and \ref{line:end_Gamma_=00007Bj,i=00007D} the wrenches that subsystem
$i$ imposes on its preceding subsystem, $p_{i}\in S$, are calculated.
The calculation of the relative poses $\dq X_{j,i}$ and wrenches
$\dq{\varGamma}_{\mathrm{tmp}}$, and the implementation of $\dq{\mathcal{W}}_{i}$
also depend on the algebraic structure representing rigid motions,
twists, and wrenches. For instance, when dual quaternion algebra is
used, wrench projections are given by (\ref{eq:vector_wrenches_s}),
whereas $\dq{\mathcal{W}}_{i}$ is obtained using Algorithm~\ref{alg:dqNE_backward_recursion}.

\textcolor{blue}{}

\begin{algorithm}
\begin{algorithmic}[1]

\Function{DMC}{$\mathcal{Q},\dot{\mathcal{Q}},\mathcal{\ddot{Q}}$}

\State{$\left(\dq X,\overline{\dq{\Xi}},\dot{\overline{\dq{\Xi}}}\right)\negthickspace\gets$\Call{DMC\_Forward\_Recursion}{$\mathcal{Q},\dot{\mathcal{Q}},\mathcal{\ddot{Q}}$}
}

\State{$\mymatrix{\Gamma}\gets$\Call{$\dq{\mathcal{N}}$}{$\dq X,\overline{\dq{\Xi}},\dot{\overline{\dq{\Xi}}}$}}

\State{\Return{$\mymatrix{\Gamma}$}}

\EndFunction

\end{algorithmic}

\caption{Dynamic modular composition (DMC) algorithm for a branched robot composed
of $s$ subsystems. \protect\label{alg:DMC}}
\end{algorithm}

\begin{algorithm}
\begin{algorithmic}[1]

\Function{DMC\_Forward\_Recursion}{$\mathcal{Q},\dot{\mathcal{Q}},\mathcal{\ddot{Q}}$}

\color{blue}

\State{\textcolor{blue}{$\dq{\Xi}_{1}\gets\left(\dq 0_{n_{1}},\dq 0_{n_{1}}\right),\dots,\dq{\Xi}_{s}\gets\left(\dq 0_{n_{s}},\dq 0_{n_{s}}\right)$}}

\State{\textcolor{blue}{$\dq X_{p_{1},1}\gets\dq 0_{n_{1}},\dots,\dq X_{p_{s},s}\gets\dq 0_{n_{s}}$}}

\color{black}

\State{$\text{queue}\gets1$}\Comment{Initialize the queue}

\State{Mark $\text{subsystem}\left(1\right)$ as visited}

\While{queue is not empty}

\State{$i\gets$ pop first element from queue}\label{line:7}

\If{$\text{subsystem}\left(i\right)$ is not black box}\label{line:begin_Xi_=00007Bi,i=00007D}

\color{blue}

\color{black}

\State{$\dq{\Xi}_{i,i}\gets$\Call{forward\_recursion}{$\myvec q_{i},\dot{\myvec q}_{i},\text{\ensuremath{\ddot{\myvec q}_{i}}}$}}\label{line:forward_recursion}

\State{$\dq{\Xi}_{i}\gets\dq{\Xi}_{i}+\dq{\Xi}_{i,i}$}\label{line:addition-of-twist-vector}

\EndIf\label{line:end_Xi_=00007Bi,i=00007D}\label{line:11}

\For{$j\gets j_{i,1}\text{\,to\,}j_{i,m_{i}}$}\label{line:begin_Xi_=00007Bpi,i=00007D}

\If{$j$ is unvisited}

\State{Push $j$ to the end of the queue}

\If{$\text{subsystem}\left(j\right)$ is not black box}

\State{\textcolor{blue}{Calculate $\myvec{\dq X}_{p_{j},j}$ using
$a_{j}$ and $\myvec q_{j}\in\mathcal{Q}$}}\label{line:calculate_X_pj_j}

\If{$\text{subsystem}\left(i\right)$ is black box}

\State{$\dq{\xi}_{0,a_{j}}^{a_{j}}\gets$\Call{get\_from\_sensors}{$i$}}

\Else

\State{Calculate $\dq{\xi}_{0,a_{j}}^{a_{j}}$ using $\dq{\Xi}_{i}$}

\EndIf

\color{blue}

\State{\textcolor{blue}{Calculate $\dq{\Xi}_{p_{j},j}$ using $\dq{\xi}_{0,a_{j}}^{a_{j}}$
and $\dq X_{p_{j},j}$}}\label{line:calculate_Xi_pj_j}

\color{black}

\State{$\dq{\Xi}_{j}\gets\dq{\Xi}_{p_{j},j}$}

\EndIf

\State{Mark $\text{subsystem}\left(j\right)$ as visited}

\EndIf

\EndFor\label{line:end_Xi_=00007Bpi,i=00007D}

\EndWhile

\State{$\left(\overline{\dq{\Xi}},\dot{\overline{\dq{\Xi}}}\right)\gets\left(\dq{\Xi}_{1},\ldots,\dq{\Xi}_{s}\right)$}

\State{$\dq X\gets\left(\dq X_{p_{1},1},\ldots,\dq X_{p_{s},s}\right)$}

\State{\Return{$\left(\dq X,\overline{\dq{\Xi}},\dot{\overline{\dq{\Xi}}}\right)$}}

\EndFunction

\end{algorithmic}

\caption{Obtain the twists and their derivatives at the $n$ CoMs of the $s$
subsystems. \protect\label{alg:DMC_FR}}
\end{algorithm}

\begin{algorithm}
\color{blue}

\begin{algorithmic}[1]

\Function{$\dq{\mathcal{N}}$}{$\dq X,\overline{\dq{\Xi}},\dot{\overline{\dq{\Xi}}}$}

\State{$\dq{\Gamma}_{1}\gets\dq 0_{n_{1}},\dots,\dq{\Gamma}_{s}\gets\dq 0_{n_{s}}$}

\For{$i\gets s\text{\,to\,}1$}

\If{$\text{subsystem}\left(i\right)$ is not black box}\label{line:begin_Wi}

\State{$\dq{\Gamma}_{i}\gets\dq{\Gamma}_{i}+\w ii$, with $\dq{\Xi}_{i}\subseteq\overline{\dq{\Xi}}\cup\dot{\overline{\dq{\Xi}}}$
}\label{line:summ_Gammai}

\EndIf\label{line:end_Wi}

\If{$\text{subsystem}\left(p_{i}\right)$ is not black box}\label{line:begin_Gamma_=00007Bj,i=00007D}

\State{Calculate $\dq X_{i,p_{i}}$ using $b_{p_{i},i}$ and $\dq X_{p_{i},i}\subseteq\dq X$}\label{line:calculate_X_i_pi}

\If{$\text{subsystem}\left(i\right)$ is black box}

\State{$\dq{\zeta}_{0,b_{p_{i},i}}^{b_{p_{i},i}}\gets$\Call{get\_from\_sensors}{$i$}}

\Else

\State{Calculate $\dq{\zeta}_{0,b_{p_{i},i}}^{b_{p_{i},i}}$ using
$\dq{\Gamma}_{i}$}

\EndIf

\State{Calculate $\dq{\Gamma}_{\mathrm{tmp}}$ using $\dq X_{i,p_{i}},\dq{\zeta}_{0,b_{p_{i},i}}^{b_{p_{i},i}}$}\label{line:calculate_Gamma_pi_due_i}

\State{$\dq{\Gamma}_{p_{i}}\gets\dq{\Gamma}_{p_{i}}+\dq{\Gamma}_{\mathrm{tmp}}$}\label{line:summ_Gamma_pi}

\EndIf\label{line:end_Gamma_=00007Bj,i=00007D}

\EndFor

\State{$\dq{\mymatrix{\Gamma}}_{\mathrm{total}}\gets\left(\dq{\Gamma}_{1},\ldots,\dq{\Gamma}_{s}\right)$}

\State{\Return{$\dq{\mymatrix{\Gamma}}_{\mathrm{total}}$}}

\EndFunction

\end{algorithmic}

\caption{Obtain the total joint wrenches of the $s$ subsystems. \protect\label{alg:N}}
\end{algorithm}

\section{\protect\label{sec:DQ-Model-Composition}Model Composition using
Dual Quaternion Algebra}

The framework presented in Section~\ref{sec:Model-Composition} is
general, has a high level of abstraction, and thus can be instantiated
into different mathematical representations. Here we present one instance
based on dual quaternion algebra,\footnote{Basic definitions are shown in Appendix~\ref{sec:Appendix-A}. For
a more comprehensive introduction of dual quaternion algebra, see
\cite{Adorno2017}.} which is particularly suitable because elements such as unit dual
quaternions and pure dual quaternions, when equipped with standard
multiplication and addition operations, form Lie groups with associated
Lie algebras \cite{Selig2005}. Furthermore, dual quaternion algebra
provides an elegant and efficient representation of homogeneous transformations
\cite{Funda1990} and screw theory \cite{Aspragathos1998}. Nonetheless,
other representations, such as the spatial algebra \cite{Featherstone2008Book}
or the Lie algebra $\mathrm{se}\left(3\right)$ \cite{Murray1994},
might also be used, as long as they capture the high-level operations
described in Section~\ref{sec:Model-Composition}.

Consider the expressions in Section~\ref{sec:Model-Composition}.
If dual quaternions are used, then $\mathsf{M}=\mathsf{T}=\mathcal{H}_{p}$,
where $\mathcal{H}_{p}$ is the set of pure dual quaternions; that
is, dual quaternions with real part equal to zero (see Appendix~\ref{sec:Appendix-A}).
Also, $\dq{\myvec 1}_{n}=\myvec 1_{n}\in\mathbb{R}^{n}$ and $\myvec{\dq 0}_{n}=\myvec 0_{n}\in\mathbb{R}^{n}$
because $0,1\in\mathbb{R}\subset\mathcal{H}$, and $0\dq x=\dq x0=0$
and $1\dq x=\dq x1=\dq x$ for any $\dq x\in\mathcal{H}$. Therefore,
the high-level expressions used to propagate twists and wrenches remain
unchanged, but the low-level expressions used to express those twists
and wrenches in different frames explicitly employ the operations
of dual quaternion algebra.

More specifically, the twist $\dq{\xi}_{0,a_{i}}^{a_{i}}\in\mathcal{H}_{p}$
at the connection point $a_{i}$ between the preceding subsystem $p_{i}$
and current subsystem $i$, propagated to each of the CoMs of subsystem
$i$, is given by the vector of twists $\overline{\dq{\Xi}}_{p_{i},i}\in\mathcal{H}_{p}^{n_{i}}$
such that
\begin{equation}
\overline{\dq{\Xi}}_{p_{i},i}=\adn{n_{i}}{\dq X_{p_{i},i}}{\dq{\xi}_{0,a_{i}}^{a_{i}}},\label{eq:vector_twists_p}
\end{equation}
where the (vector) adjoint operator $\text{Ad}_{n_{i}}$ is given
by (\ref{eq:adjoint_n}) and $\dq X_{p_{i},i}=\left[\begin{array}{ccc}
\dq x_{a_{i}}^{c_{1}} & \ldots & \dq x_{a_{i}}^{c_{n_{i}}}\end{array}\right]^{T}\in\dq{\mathcal{S}}^{n_{i}}$ is the vector of the relative poses (i.e., unit dual quaternions)
between the connection point $a_{i}$ and the $c_{1},\ldots,c_{n_{i}}$
CoMs of the $i$th subsystem. The vector of dual quaternions $\dot{\overline{\dq{\Xi}}}_{p_{i},i}\in\mathcal{H}_{p}^{n_{i}}$
is given by the time derivative of (\ref{eq:vector_twists_p}), where
each element is given by \cite{Silva2022}
\begin{multline}
\frac{d}{dt}\left(\ad{\dq x_{a_{i}}^{c_{k}}}{\dq{\xi}_{0,a_{i}}^{a_{i}}}\right)=\ad{\dq x_{a_{i}}^{c_{k}}}{\dot{\dq{\xi}}_{0,a_{i}}^{a_{i}}}\\
+\dq{\xi}_{c_{k},a_{i}}^{c_{k}}\times\left(\ad{\dq x_{a_{i}}^{c_{k}}}{\dq{\xi}_{0,a_{i}}^{a_{i}}}\right),\label{eq:vector_twists_dot_p}
\end{multline}
for $k\in\left\{ 1,\ldots,n_{i}\right\} $.

Analogously, the wrench $\dq{\zeta}_{0,b_{i,j}}^{b_{i,j}}\in\mathcal{H}_{p}$
at the connection point $b_{i,j}$ between the current subsystem $i$
and the succeeding subsystem $j$, propagated to each of the CoMs
of subsystem $i$, is given by the vector of wrenches $\mathring{\dq{\mymatrix{\Gamma}}}_{j,i}\in\mathcal{H}_{p}^{n_{i}}$
such that\footnote{It is important to notice that $\dq{\xi}_{0,a_{i}}^{a_{i}}$ in (\ref{eq:vector_twists_p})
is the twist at connection point $a_{i}$ with respect to frame $\frame 0$,
expressed in frame $\frame{a_{i}}$. Analogously, $\dq{\zeta}_{0,b_{i,j}}^{b_{i,j}}$
is the wrench at connection point $b_{i,j}$ with respect to frame
$\frame 0$, expressed in frame $\frame{b_{i,j}}$. \textcolor{blue}{We
use three indices here because the twist/wrench between two frames
can be seen from a third frame. For instance, $\dq{\xi}_{a,b}^{c}$
is the twist of frame $\frame b$ with respect to frame $\frame a$,
expressed in frame $\frame c$.}}
\begin{equation}
\mathring{\dq{\mymatrix{\Gamma}}}_{j,i}=\adn{n_{i}}{\dq X_{j,i}}{\dq{\zeta}_{0,b_{i,j}}^{b_{i,j}}},\label{eq:vector_wrenches_s}
\end{equation}
in which $\dq X_{j,i}=\left[\begin{array}{cc}
\bar{\dq X}_{j,i}^{T} & \myvec 0_{n_{i}-\eta}^{T}\end{array}\right]^{T}\in\mathcal{H}^{n_{i}}$, with $\bar{\dq X}_{j,i}=\left[\begin{array}{ccc}
\dq x_{b_{i,j}}^{0} & \ldots & \dq x_{b_{i,j}}^{\eta}\end{array}\right]^{T}\in\dq{\mathcal{S}}^{\eta}$ being the vector of relative poses (i.e., unit dual quaternions)
between the connection point $b_{i,j}$ and each of the $\eta\leq n_{i}$
joints of subsystem $i$ that precede \textbf{$b_{i,j}$}. Furthermore,
the wrench $\dq{\zeta}_{0,b_{i,j}}^{b_{i,j}}$ has the form
\[
\dq{\zeta}_{0,b_{i,j}}^{b_{i,j}}=\quat f_{0,b_{i,j}}^{b_{i,j}}+\dual\quat{\tau}_{0,b_{i,j}}^{b_{i,j}},
\]
where $\quat f_{0,b_{i,j}}^{b_{i,j}}=f_{x}\imi+f_{y}\imj+f_{z}\imk$
is the force at the connection point $b_{i,j}$ given by Newton's
second law and $\quat{\tau}_{0,b_{i,j}}^{b_{i,j}}=\tau_{x}\imi+\tau_{y}\imj+\tau_{z}\imk$
is is the torque about $b_{i,j}$ due to the change of its angular
momentum, given by the Euler's rotation equation.

Moreover, the function $\dq{\mathcal{W}}_{i}$ is given by (\ref{eq:wrench_function}),
in which the sets $\mathsf{T}$ and $\mathsf{W}$ are replaced by
the set $\mathcal{H}_{p}$, and the low-level dynamic equations of
serial kinematic chains using dual quaternions are demonstrated in
\cite{Silva2022}. Algorithms \ref{alg:dqNE}, \ref{alg:dqNE_forward_recursion},
\ref{alg:dqNE_joint_twists}, and \ref{alg:dqNE_backward_recursion}
summarize the dual quaternion Newton-Euler formalism.\footnote{For details regarding the algebraic deduction of the equations presented
in these algorithms, please refer to \cite{Silva2022}.} It is important to highlight that since these algorithms present
the \emph{low-level dynamics} of a subsystem, the indexes used in
them correspond to bodies within each subsystem.

\begin{algorithm}[h]
\begin{algorithmic}[1]

\Function{newton\_euler}{$\myvec q,\dot{\myvec q},\text{\ensuremath{\ddot{\myvec q}}}$}

\State{$\dq{\Xi}\gets$\Call{forward\_recursion}{$\myvec q,\dot{\myvec q},\text{\ensuremath{\ddot{\myvec q}}}$}
}

\State{$\mymatrix{\Gamma}\gets$\Call{backward\_recursion}{$\dq{\Xi}$}}

\State{\Return{$\mymatrix{\Gamma}$}}

\EndFunction

\end{algorithmic}

\caption{Dual Quaternion Newton-Euler Algorithm for a serial mechanism \cite{Silva2022}.
Vector $\protect\myvec{\Gamma}\in\mathcal{H}_{p}^{k}$ contains the
wrenches at each joint of the $k$-DOF serial mechanism, whereas $\protect\dq{\Xi}\in\mathcal{H}_{p}^{2k}$
is the stacked vector of twists and twists derivatives at the CoM
of each link, and $\protect\myvec q,\dot{\protect\myvec q},\ddot{\protect\myvec q}$
are the joint configurations, joint velocities, and joint accelerations,
whose dimensions depend on the number of parameters used to represent
each joint (e.g., each prismatic or revolute joint adds one dimension,
each cylindrical joint adds two, each planar, spherical, and helical
joints add three, and each 6-DoF joint adds six dimensions).\protect\label{alg:dqNE}}
\end{algorithm}

\begin{algorithm}[h]
\begin{algorithmic}[1]

\Function{forward\_recursion}{$\myvec q,\dot{\myvec q},\text{\ensuremath{\ddot{\myvec q}}}$}

\State{$\dq{\xi}_{0,c_{0}}^{c_{0}}\gets0$ and $\dot{\dq{\xi}}_{0,c_{0}}^{c_{0}}\leftarrow0$}

\For{$i\gets1\text{\,to\,}k$}

\State{$\left(\dq{\xi}_{i-1,c_{i}}^{i-1},\dot{\dq{\xi}}_{i-1,c_{i}}^{i-1}\right)\gets$\Call{joint\_twist}{$\dot{\myvec q}_{i},\text{\ensuremath{\ddot{\myvec q}_{i}}}$}}

\LineComment{Calculation of the $i$th CoM twist}

\State{$\dq{\xi}_{0,c_{i}}^{c_{i}}\gets\ad{\dq x_{c_{i-1}}^{c_{i}}}{\dq{\xi}_{0,c_{i-1}}^{c_{i-1}}}{+}\ad{\dq x_{i-1}^{c_{i}}}{\dq{\xi}_{i-1,c_{i}}^{i-1}}$}

\LineComment{Calculation of the $i$-th CoM twist derivative}

\State{$\dq{\xi}_{c_{i},c_{i-1}}^{c_{i}}\leftarrow-\ad{\dq x_{i-1}^{c_{i}}}{\dq{\xi}_{i-1,c_{i}}^{i-1}}$}

\State{$\dot{\dq{\xi}}_{0,c_{i}}^{c_{i}}\leftarrow\ad{\dq x_{c_{i-1}}^{c_{i}}}{\dot{\dq{\xi}}_{0,c_{i-1}}^{c_{i-1}}}{+}\ad{\dq x_{i-1}^{c_{i}}}{\dot{\dq{\xi}}_{i-1,c_{i}}^{i-1}}+\dq{\xi}_{c_{i},c_{i-1}}^{c_{i}}\times\left[\ad{\dq x_{c_{i-1}}^{c_{i}}}{\dq{\xi}_{0,c_{i-1}}^{c_{i-1}}}\right]$}

\EndFor

\State{$\overline{\dq{\myvec{\Xi}}}\gets\left(\dq{\xi}_{0,c_{1}}^{c_{1}},\ldots,\dq{\xi}_{0,c_{k}}^{c_{k}}\right)$,
$\dot{\overline{\dq{\myvec{\Xi}}}}\gets\left(\dot{\dq{\xi}}_{0,c_{1}}^{c_{1}},\ldots,\dot{\dq{\xi}}_{0,c_{k}}^{c_{k}}\right)$}

\State{$\dq{\Xi}\gets\left[\begin{array}{cc}
\overline{\dq{\myvec{\Xi}}}^{T} & \dot{\overline{\dq{\myvec{\Xi}}}}\end{array}^{T}\right]^{T}$}

\State{\Return{$\dq{\Xi}$}}

\EndFunction

\end{algorithmic}

\caption{Forward recursion to obtain the twists and their derivatives for the
CoM of all robot links \cite{Silva2022}. Each joint $\protect\myvec q_{i}\in\protect\myvec q=(\protect\myvec q_{1},\ldots,\protect\myvec q_{k})$
and its higher time derivatives are represented by a different number
of parameters depending on their type.\foreignlanguage{american}{\protect\label{alg:dqNE_forward_recursion}}}
\end{algorithm}

\begin{algorithm}[h]
\begin{algorithmic}[1]

\Function{joint\_twist}{$\dot{\myvec q}_{i},\text{\ensuremath{\ddot{\myvec q}}}_{i}$}

\If{revolute joint}

\State{$\dq{\xi}_{i-1,c_{i}}^{i-1}\negthickspace\leftarrow\omega_{i}\quat l_{i}^{i-1}$
and $\dot{\dq{\xi}}_{i-1,c_{i}}^{i-1}\negthickspace\leftarrow\dot{\omega}_{i}\quat l_{i}^{i-1}$}

\ElsIf{prismatic joint}

\State{$\dq{\xi}_{i-1,c_{i}}^{i-1}\negthickspace\leftarrow\dual v_{i}\quat l_{i}^{i-1}$
and $\dot{\dq{\xi}}_{i-1,c_{i}}^{i-1}\negthickspace\leftarrow\dual\dot{v}_{i}\quat l_{i}^{i-1}$}

\ElsIf{spherical joint}

\State{$\dq{\xi}_{i-1,c_{i}}^{i-1}\leftarrow\omega_{i_{x}}\imi+\omega_{i_{y}}\imj+\omega_{i_{z}}\imk$}

\State{$\dot{\dq{\xi}}_{i-1,c_{i}}^{i-1}\leftarrow\dot{\omega}_{i_{x}}\imi+\dot{\omega}_{i_{y}}\imj+\dot{\omega}_{i_{z}}\imk$}

\ElsIf{planar joint}

\State{$\dq{\xi}_{i-1,c_{i}}^{i-1}\leftarrow\omega_{i}\imk+\dual\left(v_{i_{x}}\imi+v_{i_{y}}\imj\right)$
}

\State{$\dot{\dq{\xi}}_{i-1,c_{i}}^{i-1}\leftarrow\dot{\omega}_{i}\imk+\dual\left(\dot{v}_{i_{x}}\imi+\dot{v}_{i_{y}}\imj\right)$}

\ElsIf{cylindrical joint}

\State{$\dq{\xi}_{i-1,c_{i}}^{i-1}\leftarrow\left(\omega_{i}+\dual v_{i}\right)\quat l_{i}^{i-1}$}

\State{$\dot{\dq{\xi}}_{i-1,c_{i}}^{i-1}\leftarrow\left(\dot{\omega}_{i}+\dual\dot{v}_{i}\right)\quat l_{i}^{i-1}$}

\ElsIf{helical joint}

\State{$\dq{\xi}_{i-1,c_{i}}^{i-1}\leftarrow\left(\omega_{i}+\dual h_{i}\omega_{i}\right)\quat l_{i}^{i-1}$}

\State{$\dot{\dq{\xi}}_{i-1,c_{i}}^{i-1}\leftarrow\left(\dot{\omega}_{i}+\dual h_{i}\dot{\omega}_{i}\right)\quat l_{i}^{i-1}$}

\ElsIf{6-DoF joint}

\State{$\dq{\xi}_{i-1,c_{i}}^{i-1}\negthickspace\negthickspace\negthickspace\leftarrow\omega_{i_{x}}\imi+\omega_{i_{y}}\imj+\omega_{i_{z}}\imk+\dual\left(v_{i_{x}}\imi+v_{i_{y}}\imj+v_{i_{z}}\imk\right)$}

\State{$\dot{\dq{\xi}}_{i-1,c_{i}}^{i-1}\negthickspace\negthickspace\negthickspace\leftarrow\dot{\omega}_{i_{x}}\imi+\dot{\omega}_{i_{y}}\imj+\dot{\omega}_{i_{z}}\imk+\dual\left(\dot{v}_{i_{x}}\imi+\dot{v}_{i_{y}}\imj+\dot{v}_{i_{z}}\imk\right)$}

\vspace{1mm}

\EndIf

\State{\Return{$\left(\dq{\xi}_{i-1,c_{i}}^{i-1},\dot{\dq{\xi}}_{i-1,c_{i}}^{i-1}\right)$}}

\EndFunction

\end{algorithmic}

\caption{Function to obtain the twists of some of the most commonly used joints
in robotics \cite{Silva2022}.\protect\label{alg:dqNE_joint_twists}}
\end{algorithm}

\begin{algorithm}[h]
\begin{algorithmic}[1]

\Function{backward\_recursion}{$\dq{\Xi}$}

\State{$\dq{\zeta}_{0,k+1}^{k}\leftarrow\text{external\_wrench}$}\label{line:get-external-wrench}

\For{$i\gets k\text{\,to\,}1$}

\State{$\dq{\xi}_{0,c_{i}}^{c_{i}}\gets\overline{\dq{\myvec{\Xi}}}[i]$
and $\dot{\dq{\xi}}_{0,c_{i}}^{c_{i}}\gets\dot{\overline{\dq{\myvec{\Xi}}}}[i]$}

\State{$\quat f_{0,c_{i}}^{c_{i}}\leftarrow\!m_{i}\left(\!\getd{\dot{\dq{\xi}}_{0,c_{i}}^{c_{i}}}\!{+}\!\getp{\dq{\xi}_{0,c_{i}}^{c_{i}}}\!{\times}\!\getd{\dq{\xi}_{0,c_{i}}^{c_{i}}}\!\right)$}

\State{$\quat{\tau}_{0,c_{i}}^{c_{i}}\gets\mathcal{L}_{3}\left(\quat{\mathbb{I}}_{i}^{c_{i}}\right)\getp{\dot{\dq{\xi}}_{0,c_{i}}^{c_{i}}}+\getp{\dq{\xi}_{0,c_{i}}^{c_{i}}}\times\left(\mathcal{L}_{3}\left(\quat{\mathbb{I}}_{i}^{c_{i}}\right)\getp{\dq{\xi}_{0,c_{i}}^{c_{i}}}\right)$}

\State{$\dq{\zeta}_{0,c_{i}}^{c_{i}}\leftarrow\quat f_{0,c_{i}}^{c_{i}}+\dual\quat{\tau}_{0,c_{i}}^{c_{i}}-m_{i}\quat g^{c_{i}}$}\label{line:intermediate=000020wrench}

\LineComment{Let $\myvec{\Gamma}[i]=\dq{\zeta}_{0,c_{i}}^{i-1}$}

\State{$\myvec{\Gamma}[i]\leftarrow\ad{\dq x_{c_{i}}^{i-1}}{\dq{\zeta}_{0,c_{i}}^{c_{i}}}+\ad{\dq x_{i}^{i-1}}{\dq{\zeta}_{0,c_{i+1}}^{i}}$}

\EndFor

\State{\Return{$\myvec{\Gamma}$}}

\EndFunction

\end{algorithmic}

\caption{Backward recursion to obtain the wrenches at the robot joints \cite{Silva2022}.
\protect\label{alg:dqNE_backward_recursion}}
\end{algorithm}

The following example illustrates the application of the proposed
modular composition strategy to derive the dynamic model of the subsystems
1 and 2 shown in Fig.~\ref{fig:BM_TRO2023}.
\begin{example}
\label{exa:two_subsystems_dq}Consider Example~\ref{exa:two_subsystems}
illustrated in Fig.~\ref{fig:graph_subsystems_i_and_ji}. If dual
quaternions are used, then $\myvec{\Gamma}_{1},\myvec{\Gamma}_{2}\in\mathsf{W}^{3}=\mathcal{H}_{p}^{3}$.
Also,
\[
\mathring{\dq{\mymatrix{\Gamma}}}_{2,1}=\adn 3{\dq X_{2,1}}{\dq{\zeta}_{0,b_{1,2}}^{b_{1,2}}},
\]
where $\dq{\zeta}_{0,b_{1,2}}^{b_{1,2}}$ is the wrench propagated
from subsystem 2 to subsystem 1 at the connection point $b_{1,2}$,
in which 
\begin{align*}
\dq X_{2,1} & =\left[\begin{array}{cc}
\bar{\dq X}_{2,1}^{T} & 0\end{array}\right]^{T}=\left[\begin{array}{ccc}
\dq x_{b_{1,2}}^{0} & \dq x_{b_{1,2}}^{1} & 0\end{array}\right]^{T}\in\mathcal{H}^{3}.
\end{align*}
The first two elements of $\dq X_{2,1}$ are different from zero
because point $b_{1,2}$ is connected at the second link of subsystem
1. Therefore, $\dq{\zeta}_{0,b_{1,2}}^{b_{1,2}}$ does not directly
affect its last link. Moreover,
\[
\overline{\dq{\myvec{\Xi}}}_{1,2}=\adn 3{\dq X_{1,2}}{\dq{\xi}_{0,a_{2}}^{a_{2}}},
\]
in which $\dq{\xi}_{0,a_{2}}^{a_{2}}$ is the twist propagated from
subsystem 1 to subsystem 2 at the connection point $a_{2}$, and 
\[
\dq X_{1,2}=\left[\begin{array}{ccc}
\dq x_{a_{2}}^{\breve{c}_{1}} & \dq x_{a_{2}}^{\breve{c}_{2}} & \dq x_{a_{2}}^{\breve{c}_{3}}\end{array}\right]^{T}\in\dq{\mathcal{S}}^{3},
\]
where the symbol ``~$\breve{}$~'' indicates the frames $\frame{\breve{c}_{i}}$
located at the CoM of each link in subsystem 2 (as opposed to the
frames $\frame{c_{i}}$ located at the CoM of each link in subsystem
1). Lastly,
\[
\overline{\dq{\myvec{\Xi}}}_{2,2}=\left[\begin{array}{ccc}
\dq{\xi}_{\breve{0},\breve{c}_{1}}^{\breve{c}_{1}} & \dq{\xi}_{\breve{0},\breve{c}_{2}}^{\breve{c}_{2}} & \dq{\xi}_{\breve{0},\breve{c}_{3}}^{\breve{c}_{3}}\end{array}\right]^{T}\in\mathcal{H}_{p}^{3}
\]
is the vector of twists at the CoMs of subsystem 2 that are caused
exclusively by the motion of the joints of subsystem 2.
\end{example}

\subsection{\textcolor{blue}{Computational complexity\protect\label{subsec:computational-complexity}}}

\textcolor{blue}{As presented in \cite{Silva2022}, Algorithm~\ref{alg:dqNE},
the dual quaternion Newton-Euler algorithm (dqNE), has a linear cost
in the number of DoF of a serial kinematic chain with arbitrary joints.
Therefore, each subsystem in the modular composition is calculated
with complexity $O\left(n_{i}\right)$, where $n_{i}$ is the number
of DoF of the $i$th subsystem.}

\textcolor{blue}{Algorithm~\ref{alg:DMC_FR}, which calculates the
twists and their derivatives at the $n=\sum_{i=1}^{s}n_{i}$ CoMs
of the $s$ subsystems, uses a breadth-first search algorithm to traverse
all the $s$ subsystems in the tree. In the worst case when a node
is visited, Algorithm~\ref{alg:dqNE_forward_recursion} is executed
once to calculate the forward recursion of dqNE in Line~\ref{line:forward_recursion}
with complexity $O\left(n_{i}\right)$ and Line~\ref{line:addition-of-twist-vector}
is calculated with complexity $O\left(2n_{i}\right)$ because it correspond
to a sum of vectors $\dq{\Xi}_{i},\dq{\Xi}_{i,i}\in\mathsf{T}^{2n_{i}}$
. Moreover, Line~\ref{line:7} is executed $s$ times. As such, we
have
\begin{multline*}
\underset{\text{Line }\text{\ref{line:7}}}{\underbrace{O\left(s\right)}}+\underset{\text{Line }\text{\ref{line:forward_recursion}}}{\underbrace{O\left(\sum_{i=1}^{s}n_{i}\right)}}+\underset{\text{Line }\text{\ref{line:addition-of-twist-vector}}}{\underbrace{O\left(\sum_{i=1}^{s}2n_{i}\right)}}\\
=O\left(s\right)+O\left(n\right)+2O\left(n\right)=O\left(n\right),
\end{multline*}
because $n\geq s$, as each subsystem has at least one DoF.}\footnote{\textcolor{blue}{We consider the upper bound for the Big $O$ function
\cite[p. 47]{Cormen2009book}.}}\textcolor{blue}{{} Additionally, Line~\ref{line:calculate_X_pj_j}
is executed ${s-1}$ times}\footnote{\textcolor{blue}{In the worse scenario, lines \ref{line:begin_Xi_=00007Bpi,i=00007D}
to \ref{line:end_Xi_=00007Bpi,i=00007D} are calculated for all subsystems,
except for root subsystem $n_{1}$.}}\textcolor{blue}{{} with complexity $O(n_{j})$, as we calculate the
$n_{j}$ elements of $\myvec{\dq X}_{p_{j},j}$. Furthermore, Line~\ref{line:calculate_Xi_pj_j}
is also executed $s-1$ times with complexity $O\left(n_{j}\right)$,}\footnote{\textcolor{blue}{The twist $\dq{\xi}_{0,a_{j}}^{a_{j}}$ is propagated
$2n_{j}$ times to generate the vector $\dq{\Xi}{}_{p_{j},j}\in\mathsf{T}^{2n_{j}}$.
Moreover, $O\left(2n_{j}\right)=O\left(n_{j}\right)$.}}\textcolor{blue}{{} while the remaining operations between lines \ref{line:begin_Xi_=00007Bpi,i=00007D}
to \ref{line:end_Xi_=00007Bpi,i=00007D} have constant complexity
$O\left(1\right)$ (e.g, dequeuing, reading from a sensor, etc.).
Thus, we have
\begin{multline*}
\underset{\text{Line }\ref{line:calculate_X_pj_j}}{\underbrace{O\left(\sum_{j=2}^{s}n_{j}\right)}}+\underset{\text{Line }\ref{line:calculate_Xi_pj_j}}{\underbrace{O\left(\sum_{j=2}^{s}n_{j}\right)}}+O\left(\left(s-1\right)\cdot1\right)\\
=O\left(n-n_{1}\right)+O\left(n-n_{1}\right)+O\left(s-1\right)=O\left(n\right).
\end{multline*}
 Consequently, the complexity of Algorithm~\ref{alg:DMC_FR} is $O\left(n\right)+O\left(n\right)=O\left(n\right)$,
assuming no black-box subsystems and a centralized computation.}

\textcolor{blue}{As for Algorithm~\ref{alg:N}, it traverses matrix
$\dq A$ from right to left. In the worst case, in Line~\ref{line:summ_Gammai},
it calls Algorithm~\ref{alg:dqNE_backward_recursion} $s$ times
with complexity $O\left(n_{i}\right)$ plus it performs $s$ additions
of elements in $\mathsf{W}^{n_{i}}$ with complexity $O\left(n_{i}\right)$.
Thus, 
\[
O\left(\sum_{i=1}^{s}n_{i}\right)+O\left(\sum_{i=1}^{s}n_{i}\right)=O\left(n\right)+O\left(n\right)=O\left(n\right).
\]
Additionally, between lines \ref{line:begin_Gamma_=00007Bj,i=00007D}
and \ref{line:end_Gamma_=00007Bj,i=00007D}, there are $s-1$ operations
with complexity $O\left(n_{i}\right)$ in lines \ref{line:calculate_X_i_pi},
\ref{line:calculate_Gamma_pi_due_i}, and \ref{line:summ_Gamma_pi},
and complexity $O\left(1\right)$ in the remaining lines.}\footnote{\textcolor{blue}{Only the root subsystem has no precedent subsystem.
Thus, in the worse case, lines from \ref{line:begin_Gamma_=00007Bj,i=00007D}
to \ref{line:end_Gamma_=00007Bj,i=00007D} are executed for all subsystems
but subsystem 1.}}\textcolor{blue}{{} Consequently,
\begin{align*}
\underset{\text{Lines }\ref{line:calculate_X_i_pi}\text{, }\ref{line:calculate_Gamma_pi_due_i}\text{, and }\ref{line:summ_Gamma_pi}}{\underbrace{3O\left(\!\sum_{i=2}^{s}n_{i}\!\right)}}\!+\!O\left(s-1\right) & \!=\!3O\left(n-n_{1}\right)\!+\!O\left(s-1\right)\!=\!O\left(n\right)
\end{align*}
Therefore, the total complexity of Algorithm~\ref{alg:N} is $O\left(n\right)+O\left(n\right)=O\left(n\right)$.}

\textcolor{blue}{Consequently, the total complexity of Algorithm~\ref{alg:DMC}
is $O\left(n\right)+O\left(n\right)=O\left(n\right)$. On the other
hand, the complexity of Newton-Euler-based monolithic approaches is
also $O\left(n\right)$ \cite[p. 51]{Siciliano2008}. This means that
the proposed Algorithm~\ref{alg:DMC} allows for modular composition
and the inclusion of black box subsystems without incurring a higher
complexity even in the worst case. Furthermore, it has the potential
to be faster than a monolithic solution with the aid of parallelism.
For that, there are plenty of works in the literature regarding parallel
BFS algorithms \cite{Gazit1988,Leiserson2010,Buluc2011}, which could
be applied to Algorithm~\ref{alg:DMC_FR}. Additionally, Algorithm~\ref{alg:N}
could compute the elements on the columns of matrix $\dq A$ in parallel
for each branch, although information must be synchronized when adding
wrenches at points connecting different branches. However, parallelization
is out of scope and will be explored in future work.}

\section{\protect\label{sec:Wrench-Control}Wrench-driven End-Effector Motion
Control}

Advanced general modeling techniques such as the ones in Sections~\ref{sec:Model-Composition}
and \ref{sec:DQ-Model-Composition} are valuable on their own. Nonetheless,
robot dynamic models are more useful if amenable to control design.
Therefore, this section illustrates how control laws can be easily
designed when using our proposed formalism.

The dual quaternion Newton-Euler formalism for branched robots can
be seen as the function $\dq{\mathcal{N}}\,:\,\dq{\mathcal{S}}^{n}\times\mathcal{H}_{p}^{n}\times\mathcal{H}_{p}^{n}\rightarrow\mathcal{H}_{p}^{n}$
given by
\begin{equation}
\dq{\Gamma}=\dq{\mathcal{N}}\left(\dq X,\overline{\dq{\Xi}},\dot{\overline{\dq{\Xi}}}\right)\in\mathcal{H}_{p}^{n},\label{eq:dqNE_as_a_function_of_xi_and_dxi}
\end{equation}
where $\dq{\Gamma}$ is the vector of wrenches at the $n$-DoF branched
robot joints, $\text{\ensuremath{\dq X}=\ensuremath{\begin{bmatrix}\dq x_{c_{1}}^{0},\ldots,\dq x_{c_{n}}^{0}\end{bmatrix}^{T}\in\dq{\mathcal{S}}^{n}}}$
is the vector containing the poses associated with each CoM, and $\overline{\dq{\Xi}},\dot{\overline{\dq{\Xi}}}\in\mathcal{H}_{p}^{n}$
are the stacked vector of twists and twist time derivatives at the
CoMs, respectively.

The vector $\dq{\Gamma}$ of the branched robot's joint wrenches can
be decomposed into three components, $\dq{\Gamma}=\dq{\Gamma}_{M}+\dq{\Gamma}_{C}+\dq{\Gamma}_{g}$,
where $\dq{\Gamma}_{M}=\dq{\mathcal{N}}\left(\dq X,\myvec 0_{n},\dot{\overline{\dq{\Xi}}}\right)$
is the vector of joint wrenches due to inertial components, $\dq{\Gamma}_{C}=\dq{\mathcal{N}}\left(\dq X,\overline{\dq{\Xi}},\myvec 0_{n}\right)$
is the vector of joint wrenches due to Coriolis and Centrifugal effects,
$\dq{\Gamma}_{g}=\dq{\mathcal{N}}\left(\dq X,\myvec 0_{n},\myvec 0_{n}\right)$
is the vector of joint wrenches due to gravitational effects, and
$\myvec 0_{n}\in\mathbb{R}^{n}\subset\mathcal{H}^{n}$. The function
$\dq{\mathcal{N}}$ in (\ref{eq:dqNE_as_a_function_of_xi_and_dxi})
is a natural extension from the one used for serial robots \cite{Silva2022}.

Similarly to Newton-Euler algorithms for serial kinematic chains \cite{Silva2022},
given the desired wrenches $\dq{\zeta}_{e_{l}}^{\mathcal{L}(l)}\in\mathcal{H}_{p}$
at the end-effectors of all $\ell$ leaves, where $\mathcal{L}(l)$
returns the index of the end-effector frame of the leaf subsystem
$l\in\{1,\ldots,\ell\}$, the stacked vector of external wrenches
$\dq Z_{e}=\begin{bmatrix}\dq{\zeta}_{e_{1}}^{\mathcal{L}(1)},\ldots,\dq{\zeta}_{e_{\ell}}^{\mathcal{L}(\ell)}\end{bmatrix}\in\mathcal{H}_{p}^{\ell}$
can be easily propagated during the backward recursion by letting
$\dq{\zeta}_{0,k+1}^{k}\gets\dq{\zeta}_{e_{l}}^{\mathcal{L}(l)}$
in line~\ref{line:get-external-wrench} of Algorithm~\ref{alg:dqNE_backward_recursion}
for the backward recursion in the $l$th leaf subsystem. For that,
we first define the function 
\begin{align}
\overline{\dq{\mathcal{N}}}\left(\dq X,\overline{\dq{\Xi}},\dot{\overline{\dq{\Xi}}},\dq Z_{e}\right)\, & :\,\dq{\mathcal{S}}^{n}\times\mathcal{H}_{p}^{n}\times\mathcal{H}_{p}^{n}\times\mathcal{H}_{p}^{\ell}\rightarrow\mathcal{H}_{p}^{n},\label{eq:general-newton-euler}
\end{align}
such that $\dq{\mymatrix{\Gamma}}=\overline{\dq{\mathcal{N}}}\left(\dq X,\overline{\dq{\Xi}},\dot{\overline{\dq{\Xi}}},\dq Z_{e}\right)\in\mathcal{H}_{p}^{n}$.
Similarly to (\ref{eq:dqNE_as_a_function_of_xi_and_dxi}), $\overline{\dq{\mathcal{N}}}\left(\dq X,\overline{\dq{\Xi}},\dot{\overline{\dq{\Xi}}},\dq Z_{e}\right)=\dq{\Gamma}_{M}+\dq{\Gamma}_{C}+\dq{\Gamma}_{g}+\dq{\Gamma}_{Z}$,
where $\dq{\Gamma}_{M}=\overline{\dq{\mathcal{N}}}\left(\dq X,\myvec 0_{n},\dot{\overline{\dq{\Xi}}},\myvec 0_{s}\right)$,
$\dq{\Gamma}_{C}=\overline{\dq{\mathcal{N}}}\left(\dq X,\overline{\dq{\Xi}},\myvec 0_{n},\myvec 0_{s}\right)$,
$\dq{\Gamma}_{g}=\overline{\dq{\mathcal{N}}}\left(\dq X,\myvec 0_{n},\myvec 0_{n},\myvec 0_{s}\right)$,
and $\dq{\Gamma}_{Z}=\overline{\dq{\mathcal{N}}}\left(\dq X,\myvec 0_{n},\myvec 0_{n},\dq Z_{e}\right)-\dq{\Gamma}_{g}$.

Given the vector of desired poses $\dq{\chi}_{d}=\begin{bmatrix}\dq x_{d_{1}},\ldots,\dq x_{d_{\ell}}\end{bmatrix}^{T}\in\dq{\mathcal{S}}^{\ell}$
at the $\ell$ leaves' end-effectors, the goal is to design attractive
vector fields for $\dq Z_{e}\in\mathcal{H}_{p}^{\ell}$ to ensure
that the vector $\dq{\chi}\in\dq{\mathcal{S}}^{\ell}$ of $\ell$
end-effector poses converge to $\dq{\chi}_{d}$. We design $\dq Z_{e}$
by using a straightforward extension of the controller presented in
\cite{Silva2018}. First, given the $l$th end-effector pose $\dq x_{l}$,
the desired pose $\dq x_{d_{l}}$, and the pose error \textcolor{blue}{$\tilde{\dq x}_{l}$},
the $l$th end-effector twist feedback linearizing control input is
given by
\begin{multline}
\dq U_{l}=-k_{p}\log\tilde{\dq x}_{l}-k_{v}\tilde{\dq{\xi}}_{l}^{\mathcal{L}(l)}\\
+\ad{\tilde{\dq x}_{l}^{*}}{\dot{\dq{\xi}}_{d_{l}}^{\mathcal{L}(l)}}+\left(\ad{\tilde{\dq x}_{l}^{*}}{\dq{\xi}_{d_{l}}^{\mathcal{L}(l)}}\right)\times\tilde{\dq{\xi}}_{l}^{\mathcal{L}(l)},\label{eq:U_chosen}
\end{multline}
where $\tilde{\dq{\xi}}_{l}^{\mathcal{L}(l)}\in\mathcal{H}_{p}$ is
the twist of the $l$th leaf end-effector satisfying $\dot{\tilde{\dq x}}_{l}=\frac{1}{2}\tilde{\dq x}_{l}\tilde{\dq{\xi}}_{l}^{\mathcal{L}(l)}$,
with $k_{p},k_{v}\in(0,\infty)$ being the controller gains, and $\dq{\xi}_{d_{l}}^{\mathcal{L}(i)}$
and $\dot{\dq{\xi}}_{d_{l}}^{\mathcal{L}(l)}$ are the desired $l$th
leaf end-effector twist and twist derivatives expressed in the end-effector
frame of the $l$th leaf \cite{Silva2018}. As demonstrated in \cite{Silva2018},
$\tilde{\dq x}_{l}$ converges asymptotically to $1$ when the end-effector
error dynamics is given by $\dot{\tilde{\dq x}}_{l}=\frac{1}{2}\tilde{\dq x}_{l}\dq U_{l}$
with $\dq U_{l}$ defined as in (\ref{eq:U_chosen}), which implies
that $\dq x_{l}\to\dq x_{d_{l}}$ when $t\to\infty$.

Therefore, we define $\dq Z_{e}\triangleq\begin{bmatrix}\swap{\dq U_{1}},\ldots,\swap{\dq U_{\ell}}\end{bmatrix}^{T}$
to control the $\ell$ leaves' end-effectors, where the $\mathrm{swap}$
operator defined in Appendix~\ref{sec:Appendix-A} is used to ensure
that the rotational and linear components of $\dq U_{l}$ match the
rotational and linear components of $\dq{\zeta}_{e_{l}}^{\mathcal{L}(l)}$
in $\dq Z_{e}$. Using (\ref{eq:U_chosen}) in (\ref{eq:general-newton-euler})
with $\overline{\dq{\Xi}}=\dot{\overline{\dq{\Xi}}}=\myvec 0_{n}$,
the joint wrench inputs $\dq{\Gamma}_{u}$ are given by 
\begin{align}
\dq{\Gamma}_{u} & \triangleq\overline{\dq{\mathcal{N}}}\left(\dq X,\myvec 0_{n},\myvec 0_{n},\dq Z_{e}\right)=\dq{\Gamma}_{g}+\dq{\Gamma}_{Z},\label{eq:joint-wrenches-input}
\end{align}
where $\dq{\Gamma}_{g}$ is the gravity compensation term, and $\dq{\Gamma}_{Z}$
is the vector of $n$ joint wrenches induced by the $\ell$ end-effector
wrenches $\dq Z_{e}$ that enforce the end-effector twist feedback
linearizing control inputs (\ref{eq:U_chosen}) for all $l\in\{1,\ldots,\ell\}$.
Since the inertial and Coriolis/centrifugal terms in \ref{eq:joint-wrenches-input}
are eliminated, it is equivalent to a feedback-linearizing controller
with gravity compensation. Finally, if the robot is composed of revolute
and prismatic joints, we obtain the joint forces/torques input $\myvec{\tau}_{u}\in\mathbb{R}^{n}$
by projecting $\dq{\Gamma}_{u}\in\mathcal{H}_{p}^{n}$ onto the joint
motion axes \cite{Silva2022}.

\section{\protect\label{sec:Numerical-evaluation}Numerical Evaluation and
Simulation Results}

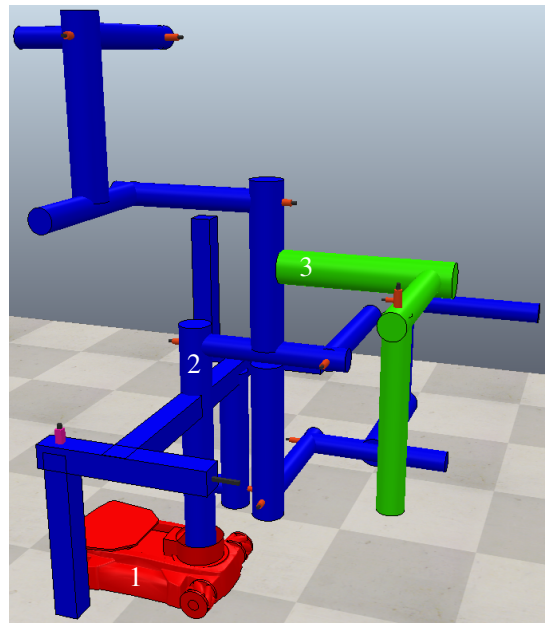
\begin{figure}[b]
\begin{centering}
{\Large\textcolor{white}{\resizebox{0.8\columnwidth}{!}{
\begingroup%
  \makeatletter%
  \providecommand\color[2][]{%
    \errmessage{(Inkscape) Color is used for the text in Inkscape, but the package 'color.sty' is not loaded}%
    \renewcommand\color[2][]{}%
  }%
  \providecommand\transparent[1]{%
    \errmessage{(Inkscape) Transparency is used (non-zero) for the text in Inkscape, but the package 'transparent.sty' is not loaded}%
    \renewcommand\transparent[1]{}%
  }%
  \providecommand\rotatebox[2]{#2}%
  \newcommand*\fsize{\dimexpr\f@size pt\relax}%
  \newcommand*\lineheight[1]{\fontsize{\fsize}{#1\fsize}\selectfont}%
  \ifx\svgwidth\undefined%
    \setlength{\unitlength}{267.13339017bp}%
    \ifx\svgscale\undefined%
      \relax%
    \else%
      \setlength{\unitlength}{\unitlength * \real{\svgscale}}%
    \fi%
  \else%
    \setlength{\unitlength}{\svgwidth}%
  \fi%
  \global\let\svgwidth\undefined%
  \global\let\svgscale\undefined%
  \makeatother%
  \begin{picture}(1,1.15327365)%
    \lineheight{1}%
    \setlength\tabcolsep{0pt}%
    \put(0,0){\includegraphics[width=\unitlength,page=1]{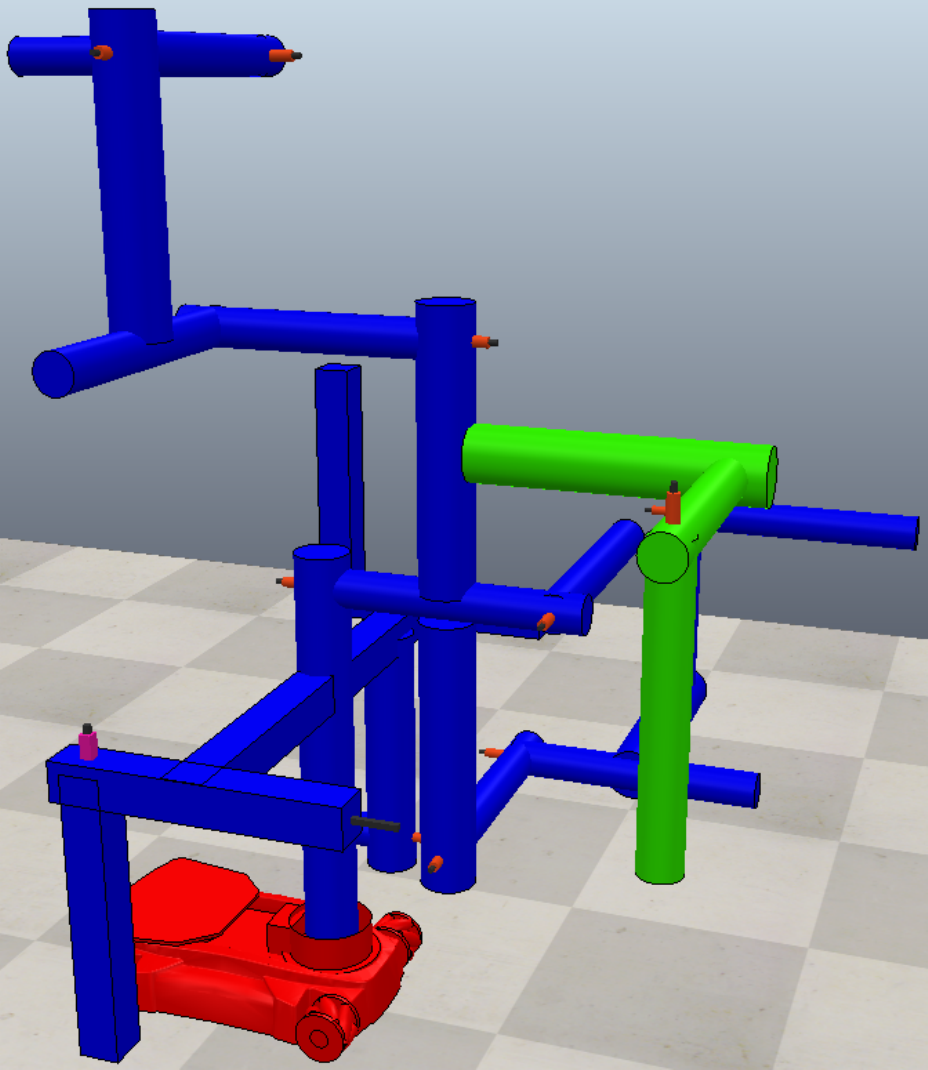}}%
    \put(0.22228459,0.06836333){\color[rgb]{1,1,1}\makebox(0,0)[lt]{\lineheight{1.25}\smash{\begin{tabular}[t]{l}1\end{tabular}}}}%
    \put(0.33175865,0.46619809){\color[rgb]{1,1,1}\makebox(0,0)[lt]{\lineheight{1.25}\smash{\begin{tabular}[t]{l}2\end{tabular}}}}%
    \put(0.54107422,0.64720582){\color[rgb]{1,1,1}\makebox(0,0)[lt]{\lineheight{1.25}\smash{\begin{tabular}[t]{l}3\end{tabular}}}}%
  \end{picture}%
\endgroup%
}}}{\Large\par}
\par\end{centering}
\caption{A $30$-DoF holonomic mobile branched manipulator (MBM) composed of
three subsystems represented by the colored rigid bodies. The second
subsystem (blue) is considered as a black box subsystem.\protect\label{fig:MBM_TRO2023}}
\end{figure}

To evaluate the accuracy and correctness of the model composition
methodology proposed in Sections~\ref{sec:Model-Composition} and
\ref{sec:DQ-Model-Composition}, we performed numerical evaluations
using two robots; namely, the fixed-base $24$-DoF branched manipulator
(BM) shown in Fig.~\ref{fig:BM_TRO2023} and the $30$-DoF holonomic
mobile branched manipulator (MBM) shown in Fig.~\ref{fig:MBM_TRO2023}.
We also include qualitative results to evaluate the wrench-driven
end-effector motion control in Section~\ref{sec:Wrench-Control}.

We implemented the simulations on the robot simulator ${\text{CoppeliaSim Edu V4.4.0}}$
\cite{Rohmer2013} with the MuJoCo \cite{Todorov2012} physics engine.
The implementation was done in MATLAB 2023b, and the computational
library DQ Robotics \cite{Adorno2021} was used for dual quaternion
algebra on a computer running Ubuntu 20.04 LTS 64 bits equipped with
an Intel i7-6500u with 8GB RAM.

\subsection{Simulation setup\protect\label{subsec:Simulation-setup}}

The BM has eight subsystems, each composed of three DoFs, some containing
prismatic or revolute joints. Therefore, the configuration vector
is defined as 
\[
\myvec q_{\mathrm{BM}}=\begin{bmatrix}\myvec q_{1}^{T},\ldots,\myvec q_{8}^{T}\end{bmatrix}^{T}\in\mathbb{R}^{24},
\]
with $\myvec q_{1},\ldots,\myvec q_{8}\in\mathbb{R}^{3}$.

Although the MBM is composed of three subsystems, the second one is
a black box in our simulation. Thus, the generalized coordinates of
the MBM were defined as
\[
\myvec q_{\mathrm{MBM}}\triangleq\left[\begin{array}{cc}
\check{\myvec q}_{\mathrm{1}}^{T} & \check{\myvec q}_{3}^{T}\end{array}\right]^{T}\in\mathbb{R}^{6},
\]
where $\check{\myvec q}_{1}\triangleq\left[\begin{array}{ccc}
x_{\mathrm{base}} & y_{\mathrm{base}} & \phi_{\mathrm{base}}\end{array}\right]^{T}\in\mathbb{R}^{3}$ is the configuration vector of subsystem 1 (i.e., the holonomic base),
with $x_{\mathrm{base}}$ and $y_{\mathrm{base}}$ being the Cartesian
coordinates and $\phi_{\mathrm{base}}\in[0,2\pi)$ being the angle
of rotation of the holonomic base. The vector $\check{\myvec q}_{3}\in\mathbb{R}^{3}$
contains the joint configurations of subsystem 3 in the MBM.

The robots followed arbitrary trajectories in the configuration space,
and their configurations ($\myvec q_{\mathrm{MBM}}$ and $\myvec q_{\mathrm{BM}}$)
and configuration velocities ($\dot{\myvec q}_{\mathrm{MBM}}$ and
$\dot{\myvec q}_{\mathrm{BM}}$) were read from CoppeliaSim. Since
the simulator does not allow the direct reading of accelerations,
$\dot{\myvec q}_{\mathrm{MBM}}$ and $\dot{\myvec q}_{\mathrm{BM}}$
were filtered using a discrete filter and used to obtain the configuration
accelerations $\ddot{\myvec q}_{\mathrm{MBM}}$ and $\ddot{\myvec q}_{\mathrm{BM}}$
by means of numerical differentiation based on Richardson extrapolation
\cite[p. 322]{Gilat2014book}. Moreover, the generalized torque vectors
$\myvec{\tau}_{\mathrm{MBM}}\in\mathbb{R}^{6}$ and $\myvec{\tau}_{\mathrm{BM}}\in\mathbb{R}^{24}$
were also read from CoppeliaSim. For the branches, this information
was directly obtained from the joints, whereas for the holonomic base
we used a force sensor at the connection point with the first link
of the second subsystem.

We used Algorithm~\ref{alg:DMC} to obtain the total wrenches, namely
$\dq{\Gamma}_{\mathrm{BM}}\in\mathcal{H}_{p}^{24}$ at the BM's joints
and $\dq{\Gamma}_{\mathrm{MBM}}\in\mathcal{H}_{p}^{4}$ at the mobile
base and joints of MBM's subsystem 3. Afterward, wrenches $\dq{\Gamma}_{\mathrm{MBM}}$
and $\dq{\Gamma}_{\mathrm{BM}}$ were projected onto the body motion
axes \cite{Silva2022} to obtain torques at revolute joints and around
the vertical axis of the mobile base, and linear forces at prismatic
joints of the branched manipulator and linear motion components of
the mobile base.

The comparisons between the generalized force waveforms obtained using
the dual quaternion Newton-Euler model composition (dqNEMC) ($\myvec{\tau}_{\mathrm{MBM}}$
and $\myvec{\tau}_{\mathrm{BM}}$) and the ones read from CoppeliaSim
were made considering \textcolor{blue}{the root mean square error
(RMSE) and} the\emph{ coefficient of multiple correlation (CMC)} \cite{Ferrari2010}
between them. The CMC provides a coefficient ranging between zero
and one that indicates how similar two given waveforms are. Identical
waveforms have CMC equal to one, whereas completely different waveforms
have CMC equal to zero. Furthermore, the CMC formulation \cite{Ferrari2010}
focuses on assessing the similarity between waveforms acquired synchronously
from different models within movement-cycles when the effect of the
model on the waveform similarity is the only variable of interest.

Moreover, we also compared our results with the recursive Newton-Euler
algorithm (sv2NE) available in Featherstone's Spatial~$v2$ package,\footnote{Available at: http://royfeatherstone.org/spatial/v2/}
a widely used and well-established library that implements the robot
dynamic modeling based on spatial algebra \cite{Featherstone2008Book}.
Spatial algebra has been used on real complex robotic platforms, such
as humanoids, thanks to its good accuracy and computational performance
\cite{Bouyarmane2019}. Therefore, it is a good basis of comparison.
However, since the Spatial~$v2$ package does not support either
mobile bases or black box subsystems, we considered the sv2NE only
for the fixed-base branched robot. For that simulation, the joint
torques $\myvec{\tau}_{\mathrm{sv2NE}}\in\mathbb{R}^{24}$ were obtained
from the sv2NE. \textcolor{blue}{Then, we calculated the RMSEs} and
CMCs between $\myvec{\tau}_{\mathrm{sv2NE}}$ and the measured joint
torques from CoppeliaSim, \textcolor{blue}{as well as the RMSEs and
CMCs between the joint torques obtained with sv2NE and the dqNEMC}.

\subsection{Model accuracy using the BM}

The BM shown in Fig.~\ref{fig:BM_TRO2023} is composed of eight subsystems.
Subsystems 1, 2, 4, 5, 6 and 8 are $3$-DoF serial kinematic chains
with revolute joints, whereas subsystems 3 and 7 are $3$-DoF serial
kinematic chains with prismatic joints. Appendix~\ref{sec:Appendix-B}
(see Table~\ref{tab:Kinematic-and-dynamic-info}) presents the kinematic
and dynamic information of those subsystems.

\textcolor{blue}{The robot joints received sinusoidal position inputs
given by $\myvec u\left(t\right)=0.01\sin\left(2\pi t\right)\myvec 1_{24}\,\unit{rad}$,
where $\myvec 1_{24}\in\mathbb{R}^{24}$ is a vector of ones. The
reference $\myvec u\left(t\right)$ was tracked by CoppeliaSim's internal
joint controllers, and the dqNEMC and the sv2NE then receive the }\textcolor{blue}{\emph{measured}}\textcolor{blue}{{}
values of $\myvec q$ and $\dot{\myvec q}$. Before being numerically
differentiated to obtain $\ddot{\myvec q}$, the joint velocities
$\dot{\myvec q}$ were filtered with a second-order discrete low-pass
Butterworth filter with normalized cutoff frequency of $100$~Hz
to filter out measurement noises introduced by CoppeliaSim.}

Table~\ref{tab:BM_TRO2023_cmc} presents the \textcolor{blue}{RMSE
and} the CMC between the joint torque waveforms obtained using dqNEMC
and sv2NE and the values obtained from CoppeliaSim, the baseline.
Both dqNEMC and sv2NE presented\textcolor{blue}{{} low mean RMSEs with
small standard deviations. Moreover,} dqNEMC and sv2NE also had mean
and minimum CMC close to one, with small standard deviation, and high
maximum CMC, thus indicating high similarity between the joint torque
waveform obtained from dqNEMC, sv2NE, and the values from CoppeliaSim.
\textcolor{blue}{Moreover, the CMCs between the dqNEMC and the sv2NE
are all equal to one, and their RMSE is of the order of $10^{-14}$.}

\begin{table*}
\textcolor{blue}{\caption{RMSE and \textcolor{black}{CMC between the joint torque waveforms
obtained through different dynamic model strategies and the values
obtained from CoppeliaSim for the the 24-DoF branched robot. The closer}
the CMC\textcolor{black}{{} is to one, the more similar the waveforms
are.\protect\label{tab:BM_TRO2023_cmc}}}
}
\centering{}%
\begin{tabular}{ccccc|cccc}
\hline 
 & \multicolumn{4}{c|}{\textcolor{blue}{RMSEs for the BM}} & \multicolumn{4}{c}{\textcolor{blue}{CMCs for the BM}}\tabularnewline
\hline 
\textcolor{blue}{Method} & \textcolor{blue}{\hspace{2mm}min\hspace{2mm}} & \textcolor{blue}{\hspace{2mm}max\hspace{2mm}} & \textcolor{blue}{\hspace{2mm}mean\hspace{2mm}} & \textcolor{blue}{\hspace{2mm}std\hspace{2mm}} & \textcolor{blue}{\hspace{2mm}min\hspace{2mm}} & \textcolor{blue}{\hspace{2mm}max\hspace{2mm}} & \textcolor{blue}{\hspace{2mm}mean\hspace{2mm}} & \textcolor{blue}{\hspace{2mm}std\hspace{2mm}}\tabularnewline
\multicolumn{1}{c}{\textcolor{blue}{dqNEMC vs. CoppeliaSim}} & \textcolor{blue}{$6.6206\times10^{-5}$} & \textcolor{blue}{$0.0364$} & \textcolor{blue}{$0.0051$} & \textcolor{blue}{$0.0081$} & \textcolor{blue}{$0.9888$} & \textcolor{blue}{$0.9999$} & \textcolor{blue}{$0.9937$} & \textcolor{blue}{$0.0037$}\tabularnewline
\multicolumn{1}{c}{\textcolor{blue}{sv2NE vs. CoppeliaSim}} & \textcolor{blue}{$6.6206\times10^{-5}$} & \textcolor{blue}{$0.0364$} & \textcolor{blue}{$0.0051$} & \textcolor{blue}{$0.0081$} & \textcolor{blue}{$0.9888$} & \textcolor{blue}{$0.9999$} & \textcolor{blue}{$0.9937$} & \textcolor{blue}{$0.0037$}\tabularnewline
\textcolor{blue}{dqNEMC vs. sv2NE} & \textcolor{blue}{$1.1041\times10^{-16}$} & \textcolor{blue}{$1.2765\times10^{-13}$} & \textcolor{blue}{$1.695\times10^{-14}$} & \textcolor{blue}{$3.1047\times10^{-14}$} & \textcolor{blue}{$1.0000$} & \textcolor{blue}{$1.0000$} & \textcolor{blue}{$1.0000$} & \textcolor{blue}{$0.0000$}\tabularnewline
\hline 
\end{tabular}
\end{table*}

Furthermore, the dqNEMC is numerically equivalent to the sv2NE, which
demonstrates the accuracy of our proposed strategy when compared to
Featherstone's spatial recursive Newton-Euler algorithm. However,
it is worth highlighting that the dqNEMC is based on a modular dynamic
model of the robot, whereas the sv2NE obtains the joint torques through
a monolithic solution (i.e., without considering the existence of
subsystems).

\begin{figure}
\begin{centering}
{\Huge\resizebox{0.93\columnwidth}{!}{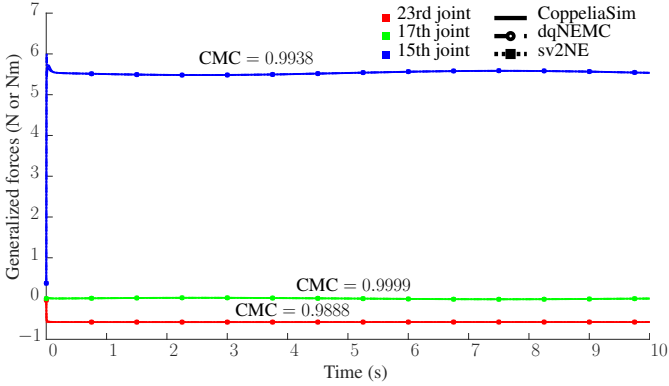}}{\Huge\par}
\par\end{centering}
\caption{\textcolor{blue}{Three joint torques/forces waveforms of the $24$-DoF
branched robot. }\textcolor{blue}{\emph{Solid}}\textcolor{blue}{{} curves
correspond to the CoppeliaSim values, whereas }\textcolor{blue}{\emph{dashed}}\textcolor{blue}{{}
curves correspond to the values obtained using the dqNEMC and }\textcolor{blue}{\emph{dot-dashed}}\textcolor{blue}{{}
curves correspond to the values obtained using the sv2NE for the joint
torque waveforms of the twenty-third ($\text{CMC}=0.9888$), seventeenth
($\text{CMC}=0.9999$), and fifteenth ($\text{CMC}=0.9938$) joints.
Because dqNEMC and sv2NE return the same results, the resulting curves
for each joint are the same.\protect\label{fig:BM_TRO2023_joint_torques}}}
\end{figure}

For qualitative analysis, Fig.~\ref{fig:BM_TRO2023_joint_torques}
presents the joint torques obtained using dqNEMC and sv2NE, alongside
the CoppeliaSim values, for the minimum, maximum, and intermediate
CMCs found during simulations. Even for the smallest value of CMC
\textcolor{blue}{(i.e., $0.9888$)}, the joint torques obtained using
our model composition formulation match closely the CoppeliaSim values.
The small discrepancies arise from discretization effects, small kinematic
and dynamic parameters uncertainties, and unmodeled effects in CoppeliaSim,
such as friction, measurement noises\textcolor{blue}{, and internal
controller dynamics}.

\subsection{Model accuracy using the MBM and black-box subsystems}

The MBM shown in Fig.~\ref{fig:MBM_TRO2023} is composed of three
subsystems, with the second being considered as a black box. The first
is a holonomic mobile base, subsystem 2 is the $24$-DoF branched
manipulator shown in Fig~\ref{fig:BM_TRO2023}, and subsystem 3 consists
of a $3$-DoF serial mechanism with prismatic joints. Appendix~\ref{sec:Appendix-B}
(see Table~\ref{tab:Kinematic-and-dynamic-info}) presents the kinematic
and dynamic information of those subsystems. Subsystems 1 and 3 do
not have access to the internal states (joint configurations, joint
velocities, and joint accelerations) of Subsystem 2, but the twists,
twists derivatives, and wrenches at the connection points are available.

\begin{figure}
\begin{centering}
\resizebox{0.4\paperwidth}{!}{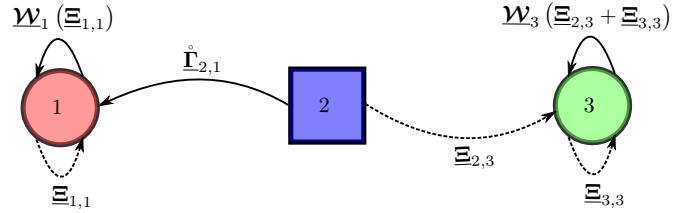}
\par\end{centering}
\caption{Graph representation of the MBM. The colored nodes follow the color
scheme adopted in Fig.~\ref{fig:MBM_TRO2023}. The black box subsystem
(subsystem 2) is represented by a square, whereas subsystems with
available internal information (subsystems 1 and 3) are represented
by circles. Despite being a black box, Subsystem 2 has outgoing edges
because twists/wrenches are available to neighboring subsystems through
sensor readings or direct communication.\protect\label{fig:MBM_TRO2023_graph}}
\end{figure}

Fig.~\ref{fig:MBM_TRO2023_graph} presents the weighted graph representing
the MBM shown in Fig.~\ref{fig:MBM_TRO2023}. Although the system
has 30 rigid bodies, the 24 rigid bodies of subsystem 2 are inside
a black box. Thus, considering only the four known rigid bodies from
subsystems 1 and 3 and the fact that we defined three subsystems but
the second is a black box, the interconnection matrix $\dq{\mymatrix A}'\in\mathcal{H}_{p}^{4\times3}$
is given by
\begin{align}
\dq{\mymatrix A}' & =\begin{bmatrix}\dq W_{1} & \mathring{\dq{\mymatrix{\Gamma}}}_{2,1} & 0\\
\myvec 0_{3} & \myvec 0_{3} & \dq W_{3}
\end{bmatrix},\label{eq:MBM=000020connection=000020matrix}
\end{align}
in which $\myvec 0_{3}\in\mathbb{R}^{3}\subset\mathcal{H}_{p}^{3}$
is a vector of zeros. Furthermore, $\dq{\mathcal{W}}_{i}=\w i{i,i}$
when $i=1$, and $\dq{\mathcal{W}}_{i}=\wadd i{p_{i},i}{i,i}$ when
$i=3$, with $p_{2}=1$ and $p_{3}=2$. Notice that, although subsystem
2 is a black box, the wrench $\dq{\zeta}_{0,b_{1,2}}^{b_{1,2}}$ at
the connection point with subsystem 1 is available through direct
measurements from a six-axis force sensor.\footnote{The information propagated by the Newton-Euler algorithm (Algorithm~\ref{alg:dqNE})
relates to joint actuation torques rather than to reaction torques
at the joints. Thus, the torque $\dq{\zeta}_{0,b_{1,2}}^{b_{1,2}}$
is the opposite of the value read from the sensor (i.e., $\dq{\zeta}_{0,b_{1,2}}^{b_{1,2}}=-\dq{\zeta}_{\mathrm{sensor}}$).}\textcolor{blue}{{} }The twist $\dq{\xi}_{0,p_{3}}^{p_{3}}$ and twist
derivative $\dot{\dq{\xi}}_{0,p_{3}}^{p_{3}}$at the connection point
between subsystems 2 and 3, which are necessary to calculate $\dq{\Xi}_{p_{3},3}$
in $\dq W_{3}$, were obtained directly from CoppeliaSim for simplicity.
Nonetheless, this information could be either measured through appropriate
sensors or communicated by subsystem 2 after its internal calculations.\footnote{Because subsystem 2 is a black box, its internal states cannot be
accessed. Nonetheless, the connection points can be regarded as the
outputs of black-box subsystems.} Therefore, $\mathring{\dq{\mymatrix{\Gamma}}}_{2,1}$ is calculated
using (\ref{eq:vector_wrenches_s}), and $\dq{\Xi}_{p_{3},3}$ is
calculated using (\ref{eq:vector_twists_p}) and (\ref{eq:vector_twists_dot_p})
, where $\dq X_{2,1}$ and $\dq X_{p_{3},3}$ used in those calculations
only require information from subsystem 1 and subsystem 3, respectively.

The robot followed sinusoidal joint/base \textcolor{blue}{reference
position} trajectories given by \textcolor{blue}{$\myvec u\left(t\right)=0.01\sin\left(2\pi t\right)\myvec 1_{30}\,\unit{rad}$},
which were \textcolor{blue}{tracked by CoppeliaSim's internal joint/base
controllers}. Notice that the joints of the black-box subsystem 2
are actuated only to simulate possible internal dynamics. However,
subsystems 1 and 3 do not have access to this information or the internal
states of subsystem 2. Before being used in the model, information
obtained from CoppeliaSim ($\dq{\zeta}_{0,b_{1,2}}^{b_{1,2}}$, $\dq{\xi}_{0,p_{3}}^{p_{3}}$
and $\dot{\myvec q}$) was filtered with a second-order discrete low-pass
Butterworth filter with normalized cutoff frequency of \textcolor{blue}{$30$~Hz}.
Similarly to the joint accelerations, the twist time derivative $\dot{\dq{\xi}}_{0,p_{3}}^{p_{3}}$
was obtained using numerical differentiation based on Richardson extrapolation.

As explained in Section~\ref{subsec:Simulation-setup}, the dqNEMC
then receives the values of $\myvec q$, $\dot{\myvec q}$, and $\ddot{\myvec q}$.
The comparison is made considering the generalized forces $\myvec{\tau}$
read from the joints and the force sensor at the base.

Table~\ref{tab:cmc-BMB_TRO2023} presents \textcolor{blue}{the RMSE
and} the CMC between the joint torque waveforms obtained through the
dqNEMC and the values obtained from CoppeliaSim. As with the previous
simulation, the dqNEMC presented \textcolor{blue}{low RMSEs and} mean
and minimum CMC close to one, with small standard deviation, and high
maximum CMC, thus indicating high similarity between the joint torque
waveform obtained from dqNEMC and the values from CoppeliaSim.

\begin{table*}
\caption{RMSE and CMC between the joint torque waveforms obtained through the
dqNEMC and the values obtained from CoppeliaSim for the $30$-DoF
holonomic branched robot. The closer the CMC is to one, the more similar
the waveforms are.\protect\label{tab:cmc-BMB_TRO2023}}

\centering{}%
\begin{tabular}{ccccc|cccc}
\hline 
 & \multicolumn{4}{c|}{\textcolor{blue}{RMSEs for the MBM}} & \multicolumn{4}{c}{\textcolor{blue}{\hspace{1mm}CMCs for the MBM}}\tabularnewline
\hline 
\textcolor{blue}{Method} & \textcolor{blue}{\hspace{2mm}min\hspace{2mm}} & \textcolor{blue}{\hspace{2mm}max\hspace{2mm}} & \textcolor{blue}{\hspace{2mm}mean\hspace{2mm}} & \textcolor{blue}{\hspace{2mm}std\hspace{2mm}} & \textcolor{blue}{\hspace{2mm}min\hspace{2mm}} & \textcolor{blue}{\hspace{2mm}max\hspace{2mm}} & \textcolor{blue}{\hspace{2mm}mean\hspace{2mm}} & \textcolor{blue}{\hspace{2mm}std\hspace{2mm}}\tabularnewline
\multicolumn{1}{c}{\textcolor{blue}{dqNEMC}} & \textcolor{blue}{$0.0004$} & \textcolor{blue}{$0.3917$} & \textcolor{blue}{$0.0697$} & \textcolor{blue}{$0.1578$} & \textcolor{blue}{$0.9498$} & \textcolor{blue}{$1.0000$} & \textcolor{blue}{$0.9856$} & \textcolor{blue}{$0.0199$}\tabularnewline
\hline 
\end{tabular}
\end{table*}

\begin{figure}[h]
\begin{centering}
{\Huge\resizebox{0.98\columnwidth}{!}{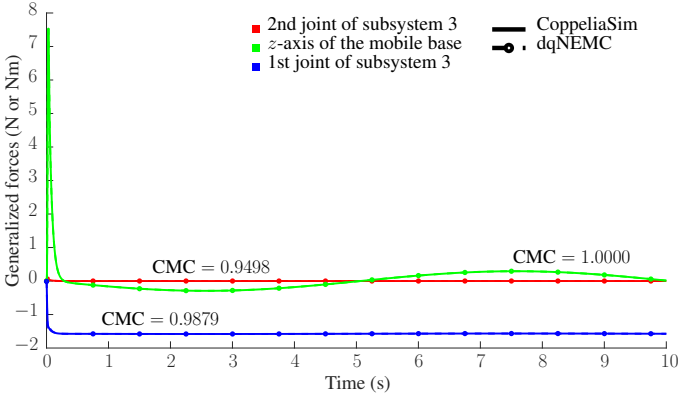}}{\Huge\par}
\par\end{centering}
\caption{\textcolor{blue}{Torque waveforms for three joints of the MBM. Solid
curves correspond to the CoppeliaSim values, whereas dashed curves
with circle markers correspond to the values obtained using the dqNEMC
for the joint torque waveforms of the first ($\text{CMC}=0.9879$)
and second ($\text{CMC}=0.9498$) joints of subsystem 3 and the generalized
force along the $z$-axis of subsystem 1 ($\text{CMC}=1.0000$).\protect\label{fig:MBM_TRO2023_joint_torques}}}
\end{figure}

For qualitative analysis, Fig.~\ref{fig:MBM_TRO2023_joint_torques}
presents the joint torques obtained using dqNEMC alongside the CoppeliaSim
values, for the minimum, maximum, and intermediate CMCs found during
simulations. Even for the smallest value of CMC \textcolor{blue}{(i.e.,
$0.9498$)}, the joint torques obtained using our model composition
formulation match closely the CoppeliaSim values. The small discrepancies
arise from discretization, unmodeled effects in CoppeliaSim, and small
uncertainties in the kinematic and dynamic models.

\subsection{Closed-loop control of the MBM\textcolor{blue}{\protect\label{subsec:MBM-Control}}}

Consider the MBM shown in Fig.~\ref{fig:MBM_TRO2023}. However, rather
than considering the whole BM as a black box subsystem, we consider
all its eight subsystems explicitly as we want to control the robot.
As a result, the MBM consists of 10 subsystems in this simulation.
Subsystem 1 in Fig~\ref{fig:MBM_TRO2023} is the root node and is
attached to subsystem 1 from the BM, which is now labeled subsystem
2. Consequently, all subsystem labels shown in Fig.~\ref{fig:BM_TRO2023}
are increased by one. Finally, subsystem 10 is attached to the first
link of the new subsystem 3 (previously labeled subsystem 2 in Fig.~\ref{fig:BM_TRO2023}).

Using (\ref{eq:joint-wrenches-input}), we control the end-effector
pose of the $\ell$ leaf subsystems of the MBM. Subsystems 4, 5, 9,
and 10 received desired end-effector poses within their workspace,
whereas desired pose of subsystem 7 was given by its initial end-effector
pose.

\begin{figure}
\begin{centering}
{\Huge\resizebox{0.98\columnwidth}{!}{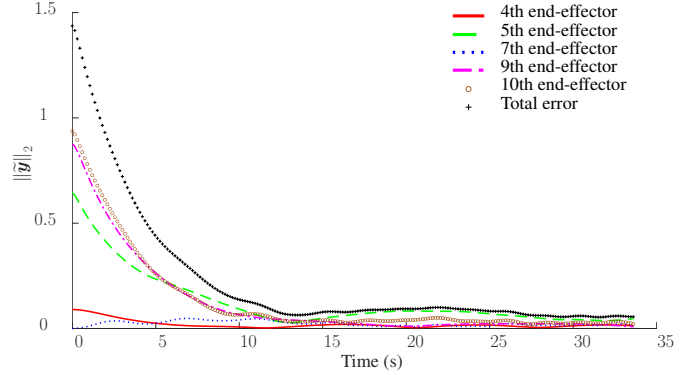}}{\Huge\par}
\par\end{centering}
\textcolor{blue}{\caption{Norm of the closed loop error response for the end-effectors of the
$\ell$ leaf subsystems of the MBM. The \emph{solid red} curve, the
\emph{dashed green} curve, the \emph{blue dotted} curve, the \emph{magenta
dash-dotted} curve, and the \emph{circled brown} curve correspond,
respectively, to the end-effectors of subsystems 4, 5, 7, 9, and 10,
whereas the \emph{crossed black} curve corresponds to the norm of
the total error (\ref{eq:total_error}).\protect\label{fig:norm-error-MBM}}
}
\end{figure}

Fig.~\ref{fig:norm-error-MBM} shows the norm of the pose error for
all leaf subsystems' end-effectors, each given by $\norm{\tilde{\myvec y}_{l}}_{2}$
with $\tilde{\myvec y}_{l}=\vector_{6}\left(\log\tilde{\dq x}_{l}\right)$
for all $l\in\{4,5,7,9,10\}$, and the norm of the total error given
by $\norm{\tilde{\myvec y}}_{2}$ where 
\begin{align}
\tilde{\myvec y} & =\begin{bmatrix}\tilde{\myvec y}_{4}^{T} & \tilde{\myvec y}_{5}^{T} & \tilde{\myvec y}_{7}^{T} & \tilde{\myvec y}_{9}^{T} & \tilde{\myvec y}_{10}^{T}\end{bmatrix}^{T}.\label{eq:total_error}
\end{align}
The oscillatory behavior of the error response is expected because
the desired closed-loop error dynamics for the end-effector poses
due to control law (\ref{eq:U_chosen}) is described by a second-order
system \cite{Silva2018}. Nonetheless, the error for the end-effector
poses of all subsystems decay and achieve steady-state.

\section{\protect\label{sec:Conclusions}Conclusions}

This paper has presented a modular composition strategy for the dynamic
modeling of branched robots that provides a high level of abstraction
and enables combining the dynamics of simpler mechanisms to obtain
the whole-body dynamics. The proposed formalism requires only twists,
twist time derivatives, and wrenches at the connection points between
different subsystems to find the coupled dynamics of the combined
mechanisms. Thus, distinct from other approaches in the literature,
our strategy works even when subsystems are black boxes, as long as
the required information at the connection points is known, which
can be done through sensor readings. This property is particularly
appealing in modular robotics, where different independent kinematic
structures can be arbitrarily combined, making the preprogramming
of dynamic modeling equations for the whole system impractical. Furthermore,
we also proposed a graph representation for the complete branched
mechanism, where each vertex is an open kinematic chain, and the wrenches
at the joints result from the graph interconnection matrix. This representation
enables obtaining the dynamics of the whole system through straightforward
algebraic operations. \textcolor{blue}{Moreover, all these features
are achieved, in the worst case, with the same linear complexity in
the number of DoFs as well-established monolithic recursive Newton-Euler
algorithms.  }Additionally, we have presented a formulation for wrench-driven
end-effector motion control to illustrate the applicability of the
model obtained through the recursive equations of the dynamic model
decomposition.

Simulation results have shown that our strategy is numerically equivalent
to monolithic solutions, such as Featherstone's Spatial~$v2$ Newton-Euler
algorithm, whose dynamic model is built considering the whole open
kinematic tree at once and assumes full knowledge of twists, twists
derivatives, and wrenches acting on all rigid bodies in the multibody
system. Those results have also shown that our formalism can be effectively
used to obtain the dynamic model of a branched mobile manipulator
containing a large black box subsystem. Indeed, the values of joint
torques in non-black systems closely matched the ones given by the
simulator, as attested by high CMCs, with small discrepancies arising
solely due to unmodeled effects (e.g., measurement noise in CoppeliaSim
and discretization effects). Furthermore, the numerical results for
the motion control of the end-effectors of a robot composed of a branched
kinematic structure on top of a holonomic base show that our model
can be used for control design.

Future work will focus on extending the proposed formulation to closed
kinematic chains and exploiting parallelization techniques for real-time
calculation of the dynamic model of complex branched robots with a
very large number, potentially thousands of degrees of freedom. We
intend to explore applications in self-reconfiguring robots, in which
the method must handle the inclusion or removal of possible unknown
modules at execution time. Lastly, we will also expand the control
formulation to handle conflicting desired wrenches and geometric constraints
across the kinematic tree.

\appendices{}

\section{Dual Quaternion Algebra\protect\label{sec:Appendix-A}}

Dual quaternions \cite{Selig2005} are elements of the set
\begin{align*}
\mathcal{H} & \triangleq\{\quat h_{\mathcal{P}}+\dual\quat h_{\mathcal{D}}:\quat h_{\mathcal{P}},\quat h_{\mathcal{D}}\in\mathbb{H},\,\dual\neq0,\,\dual^{2}=0\},
\end{align*}
where $\mathbb{H}\triangleq\{h_{1}+\imi h_{2}+\imj h_{3}+\imk h_{4}\::\:h_{1},h_{2},h_{3},h_{4}\in\mathbb{R}\}$
is the set of quaternions, in which $\imi$, $\imj$ and $\imk$ are
imaginary units with the properties $\imi^{2}=\imj^{2}=\imk^{2}=\imi\imj\imk=-1$
\cite{Selig2005}. Addition and multiplication of dual quaternions
are analogous to their counterparts of real and complex numbers. One
must only respect the properties of the dual unit $\dual$ and imaginary
units $\imi,\imj,\imk$. Given $\dq h\in\mathcal{H}$, with $\dq h=\quat h_{\mathcal{P}}+\dual\quat h_{\mathcal{D}}$,
we define $\mathrm{swap}:\mathcal{H}\to\mathcal{H}$ such that $\swap{\dq h}\triangleq\quat h_{\mathcal{D}}+\dual\quat h_{\mathcal{P}}$.

The subset $\dq{\mathcal{S}}=\left\{ \dq h\in\mathcal{H}:\norm{\dq h}=1\right\} $
of unit dual quaternions, where $\norm{\dq h}=\sqrt{\dq h\dq h^{*}}=\sqrt{\dq h^{*}\dq h}$,
with $\dq h^{*}$ being the conjugate of $\dq h$ \cite{Adorno2017},
is used to represent poses (position and orientation) in the three-dimensional
space and form the group $\spinr$ under the multiplication operation.
Any $\dq x\in\dq{\mathcal{S}}$ can always be written as $\dq x=\quat r+\dual\left(1/2\right)\quat p\quat r$,
where $\quat p=\imi x+\imj y+\imk z$ represents the position $\left(x,y,z\right)$
and $\quat r=\cos\left(\phi/2\right)+\quat n\sin\left(\phi/2\right)$
represents a rotation, in which $\phi\in[0,2\pi)$ is the rotation
angle around the rotation axis $\quat n\in\mathbb{H}_{p}\cap\mathbb{S}^{3}$,
with $\mathbb{H}_{p}\triangleq\left\{ \quat h\in\mathbb{H}:\real{\quat h}=0\right\} $,
where $\real{h_{1}+\imi h_{2}+\imj h_{3}+\imk h_{4}}\triangleq h_{1}$,
and $\mathbb{S}^{3}=\left\{ \quat h\in\mathbb{H}:\norm{\quat h}=1\right\} $
\cite{Selig2005}. Given $\dq x\in\dq{\mathcal{S}}$, the logarithmic
mapping is defined as $\log\dq x\triangleq\left(\phi\quat n+\dual\quat p\right)/2.$

The set $\mathcal{H}_{p}=\left\{ \dq h\in\mathcal{H}:\real{\dq h}=0\right\} $
of pure dual quaternions is used to represent twists and wrenches,
which are represented in different coordinate systems using the adjoint
operator $\mathrm{Ad}:\dq{\mathcal{S}}\times\mathcal{H}_{p}\to\mathcal{H}_{p}$.
For instance, consider the twist $\dq{\xi}^{a}\in\mathcal{H}_{p}$
expressed in frame $\frame a$ and the unit dual quaternion $\dq x_{a}^{b}=\quat r_{a}^{b}+\dual\frac{1}{2}\quat r_{a}^{b}\quat p_{b,a}^{a}$
that represents the rigid motion from $\frame b$ to $\frame a$.
The same twist is expressed in frame $\frame b$ as
\begin{equation}
\dq{\xi}^{b}=\ad{\dq x_{a}^{b}}{\dq{\xi}^{a}}=\dq x_{a}^{b}\dq{\xi}^{a}\left(\dq x_{a}^{b}\right)^{*},\label{eq:adj_transf}
\end{equation}
where $\dq{\xi}^{a}=\quat{\omega}_{b,a}^{a}+\dual\dot{\quat p}_{b,a}^{a}$
and $\dq{\xi}^{b}=\quat{\omega}_{b,a}^{b}+\dual\left(\dot{\quat p}_{b,a}^{b}+\quat p_{b,a}^{b}\times\quat{\omega}_{b,a}^{b}\right)$,
and $\quat{\omega}_{b,a}^{a},\quat{\omega}_{b,a}^{b}\in\mathbb{H}_{p}$
are the angular velocities that satisfy $\dot{\quat r}_{a}^{b}=\frac{1}{2}\quat{\omega}_{b,a}^{b}\quat r_{a}^{b}=\frac{1}{2}\quat r_{a}^{b}\quat{\omega}_{b,a}^{a}$
\cite{Adorno2017}. Wrenches are represented analogously but with
the linear force in the primary part and the moment in the dual part
\cite{Adorno2011}. For instance, given the wrench $\dq{\zeta}^{a}=\quat f^{a}+\dual\quat m^{a}$
in frame $\frame a$, with $\quat f^{a},\quat m^{a}\in\mathbb{H}_{p}$,
it can be expressed in frame $\frame b$ through the transformation
$\dq{\zeta}^{b}=\ad{\dq x_{a}^{b}}{\dq{\zeta}^{a}}$. Given $\dq h_{p}\in\mathcal{H}_{p}$,
such that $\dq h_{p}=h_{1}\imi+h_{2}\imj+h_{3}\imk+\dual\left(h_{4}\imi+h_{5}\imj+h_{6}\imk\right)$,
it can be mapped to $\mathbb{R}^{6}$ by using $\vector_{6}\left(\dq h_{p}\right)\triangleq\begin{bmatrix}h_{1} & h_{2} & h_{3} & h_{4} & h_{5} & h_{6}\end{bmatrix}^{T}$.

The next definition extends the adjoint operation (\ref{eq:adj_transf})
to the set $\mathcal{H}_{p}^{n}$.
\begin{defn}
\label{def:adjoint_n}Given a vector of poses $\dq X=\left[\begin{array}{cccc}
\dq x_{1} & \dq x_{2} & \cdots & \dq x_{n}\end{array}\right]^{T}\in\dq{\mathcal{S}}^{n}$ and a pure dual quaternion $\dq a\in\mathcal{H}_{p}$, the operator
$\text{Ad}_{n}:\dq{\mathcal{S}}^{n}\times\mathcal{H}_{p}\rightarrow\mathcal{H}_{p}^{n}$
is defined as
\begin{align}
\adn n{\dq X}{\dq a} & \triangleq\textrm{diag}\left(\dq X\right)\dq a\dq X^{*T}\nonumber \\
 & =\begin{bmatrix}\ad{\dq x_{1}}{\dq a} & \ad{\dq x_{2}}{\dq a} & \cdots & \ad{\dq x_{n}}{\dq a}\end{bmatrix}^{T}\label{eq:adjoint_n}
\end{align}
where $\textrm{diag}\left(\dq X\right)\in\mathcal{H}_{p}^{n\times n}$
is a diagonal matrix with the elements of $\dq X$ in the main diagonal,
and $\dq X^{*}$ is the conjugate transpose of $\dq X$.
\end{defn}

\section{Robot Parameters\protect\label{sec:Appendix-B}}

Table~\ref{tab:Kinematic-and-dynamic-info} presents the kinematic
and dynamic parameters of the $3$-DoF manipulators with revolute/prismatic
joints and the holonomic mobile base used in the branched robots shown
in Figs.~\ref{fig:BM_TRO2023} and \ref{fig:MBM_TRO2023}.

{\renewcommand{\arraystretch}{1.2}

\begin{table*}
\caption{Kinematic and dynamic information of the robots used in simulation.\protect\label{tab:Kinematic-and-dynamic-info}}

\centering{}%
\begin{tabular}{cccccccl}
\hline 
\multirow{2}{*}{Link} & \multicolumn{4}{c}{D-H parameters} & \multirow{2}{*}{CoM} & \multirow{2}{*}{Mass (Kg)} & \multirow{2}{*}{Inertia tensor ($\quat{\mathbb{I}}\triangleq\left(\quat i_{x},\quat i_{y},\quat i_{z}\right)\in\mathbb{H}_{p}^{3}$)}\tabularnewline
 & $a_{i}$ & $\alpha_{i}$ & $d_{i}$ & $\theta_{i}$ &  &  & \tabularnewline
\hline 
\hline 
\multicolumn{8}{c}{\textbf{3-DoF manipulators with revolute/prismatic joints}}\tabularnewline
1 & $0$ & $\frac{\pi}{2}$ & $0.187$ & $\frac{\pi}{2}$ & $-0.187\imj$ & $0.8$ & \foreignlanguage{american}{$\quat i_{x}=0.80\imi$, $\quat i_{y}=0.80\imj$, $\quat i_{z}=0.80\imk$\foreignlanguage{english}{\allowbreak}}\tabularnewline
\cline{8-8}
2 & $0$ & $\frac{\pi}{2}$ & $0.43$ & $\frac{\pi}{2}$ & $-0.195\imj$ & $0.5$ & \foreignlanguage{american}{$\quat i_{x}=0.50\imi$, $\quat i_{y}=0.50\imj$, $\quat i_{z}=0.50\imk$\foreignlanguage{english}{\allowbreak}}\tabularnewline
\cline{8-8}
3 & $0$ & $0$ & $0$ & $0$ & $0.235\imk$ & $0.10$ & \foreignlanguage{american}{$\quat i_{x}=0.10\imi$\foreignlanguage{english}{, }$\quat i_{y}=0.10\imj$\foreignlanguage{english}{,
}$\quat i_{z}=0.10\imk$\foreignlanguage{english}{\allowbreak}}\tabularnewline
\hline 
\multicolumn{8}{c}{\textbf{Holonomic mobile base}}\tabularnewline
N/A & N/A & N/A & N/A & N/A & $0$ & $80$ & \foreignlanguage{american}{$\quat i_{x}=40\imi$\foreignlanguage{english}{, }$\quat i_{y}=40\imj$\foreignlanguage{english}{,
}$\quat i_{z}=40\imk$\foreignlanguage{english}{\allowbreak}}\tabularnewline
\hline 
\end{tabular}
\end{table*}

\bibliographystyle{IEEEtran}
\bibliography{bibliography}

\end{document}